\documentclass[twoside,11pt]{article}

\usepackage{jmlr2e}

\usepackage{microtype}
\usepackage{textgreek}
\usepackage{algorithm,algorithmic}
\usepackage{eqparbox}

\usepackage{graphicx}
\usepackage{capt-of}% or \usepackage{caption}
\usepackage{varwidth}
\usepackage{pifont}
\usepackage[dvipsnames]{xcolor}
\usepackage{makecell}
\usepackage{subfigure}
\usepackage{amsmath}
\usepackage{mathtools}
\usepackage{eucal}
\usepackage{bbm}
\usepackage{color}
\usepackage{xcolor}
\usepackage{multirow}
\usepackage{booktabs} % for professional tables
\usepackage{amsfonts}
\newcommand{\argmax}{\operatornamewithlimits{argmax}}
\newcommand{\argmin}{\operatornamewithlimits{argmin}}

\usepackage{hyperref}

\jmlrheading{24}{2023}{1-39}{8/09 - Revised 07/16}{09/06}{21-0899}{Dat Hong,  Tong Wang and Stephen Baek}

\ShortHeadings{ProtoryNet}{Hong, Wang and Baek}
\firstpageno{1}

\begin{document}

\title{Interpretable Text Classification Via Prototype Trajectories}

    \author{\name Dat Hong  \email dat.hong@yale.edu \\
        \addr  School of Management\\Yale University  \\ New Haven, CT, USA        \AND  \name Tong Wang\thanks{Corresponding author: Tong Wang.}  \email tong.wang.tw687@yale.edu \\
        \addr School of Management\\Yale University \\ New Haven, CT, USA \AND \name Stephen Baek  \email baek@virginia.edu \\
        \addr  School of Data Science\\University of Virginia  \\ Charlottesville, VA, USA
    }
\editor{Qiaozhu Mei}
\maketitle
\begin{abstract}
We propose a novel interpretable deep neural network for text classification, called ProtoryNet, based on a new concept of prototype trajectories. Motivated by the prototype theory in modern linguistics, ProtoryNet makes a prediction by finding the most similar prototype for each sentence in a text sequence and feeding an RNN backbone with the proximity of each sentence to the corresponding active prototype. The RNN backbone then captures the temporal pattern of the prototypes, which we refer to as \textit{prototype trajectories}. Prototype trajectories enable intuitive and fine-grained interpretation of the reasoning process of the RNN model, in resemblance to how humans analyze texts. We also design a prototype pruning procedure to reduce the total number of prototypes used by the model for better interpretability. Experiments on multiple public datasets demonstrate that ProtoryNet achieves higher accuracy than the baseline prototype-based deep neural net and narrows the performance gap when compared to state-of-the-art black-box models. In addition, after prototype pruning, the resulting ProtoryNet models only need less than or around \textcolor{black}{20 prototypes for all datasets}, which significantly benefits interpretability. Furthermore, we report survey results indicating that human users find ProtoryNet more intuitive and easier to understand compared to other prototype-based methods.
\end{abstract}

\section{Introduction}
Deep neural networks have become widely adopted for numerous tasks involving unstructured data, such as texts. State-of-the-art deep neural networks for text data include recurrent neural network based models with attention mechanism \citep{wang2016attention,galassi2020attention},  convolutional neural networks \citep{yin2017comparative,young2018recent}, or Transformers \citep{devlin2018bert,liu2019roberta}. Despite achieving good predictive performance, there is a growing demand for AI models in real-world applications to be interpretable. This allows end-users to understand the decision-making process and establish trust, facilitating their adoption and collaboration with the model. However, in their conventional form, deep neural networks are black-boxes, where features undergo multiple layers of non-linear transformation, which quickly become intractable and incomprehensible to users. %Most of the state-of-the-art DNN are based on recurrent neural networks (RNN) have been widely adopted in natural language processing. RNNs achieve the state-of-the-art performance by utilizing the contextual information in a ``memory'' mechanism modeled via hidden/cell states. Albeit the benefit, however, the memory mechanism obstructs the interpretation of model decisions: as hidden states are carried over time, various pieces of information get intertwined across time steps, making RNN models a ``black box'' inherently.

The black-box nature of existing DNNs for text data has motivated a body of research  focused on achieving model \textit{interpretability}. This research can be broadly categorized into two directions. One popular group of approaches is to generate \emph{post-hoc} explanations \citep{jacovi2018understanding}. However, they generally suffer from fundamental limitations in providing post-hoc explanations. As pointed out by recent research \citep{rudin2019stop, alvarez2018robustness}, there may exist inconsistency and unfaithfulness in the explanations, since explainer methods only try to approximate the decision-making process, but they are not the real decision-maker. Another type of approach to understanding the inner workings in deep neural networks is to leverage certain architecture designs, such as \textit{attention-based} methods. %As will be discussed in Section~\ref{sec:related}, t
The attention-based approaches \citep{karpathy2015visualizing, strobelt2017lstmvis, choi2016retain, guo2018interpretable}  weigh the importance of each hidden state in a sequence. However, while a few of them could be expository, the attention weights are, in general, not always intelligible, as pointed out by \citet{jain2019attention}. 
Furthermore, analyzing attention weights necessitates a certain level of comprehension of RNN functioning in theory. Hence, novice/non-technical users may find it difficult to understand, and, thus, the broader use in real-world applications might not be so feasible.

Recent efforts have been invested in redesigning neural networks towards making them \emph{inherently interpretable}, based on the classic framework of prototypical learning \citep{datta1995learning}. 
These models utilize prototypes to provide intuitive explanations for decisions, with each prototype representing a typical case from past observations. This process parallels how human experts, such as doctors or judges, make decisions by referring to similar previous cases and drawing conclusions from them. %For for a given sequence, a prototype-based approach looks up a few representative examples, or \textit{prototypes}, from the training data set and deduces a decision based on them.
%Here, in the context of text classification, an input text, a review, for instance, is predicted to be positive because it is similar to some other reviews from the training data who are also positive. 
From the interpretability standpoint, such prototypes provide an intuitive explanation of how the model has reached a conclusion in a form that even a layperson can understand, as long as they understand the similarity by reading the prototypes.  Various existing prototype-based models adopt this reasoning logic  \citep{chen2019looks, ming2019interpretable,arik2020protoattend}. For instance, ProSeNet  \citep{ming2019interpretable} predicts the positivity of a review by comparing it to other positive reviews in the training data, where the final score is the sum of contributions from these prototypes. 

In this paper, we identify two designs in existing prototype-based models that are not so suitable for text data. First, existing prototype-based DNN models define prototypes at the document level \citep{ming2019interpretable,arik2020protoattend} and decompose a prediction into contributions from each prototype. However, when the input text is long, it becomes difficult to relate the input document to prototypes given the possible complexity of the document, which may include twists, changes of tones, etc. For example, if the input is as simple as ``The food is very delicious!'', it is easy for a user to understand why it is similar to the prototype ``Great food!''. But if the input consists of 10 sentences that first talks about the long wait at the restaurant, and proceeds to compliment the food, but then complains about the rude waiter, and finally concludes that the overall experience was not worth the money spent, it is then difficult for a user to understand why this input is similar to a particular set of prototypes that also talk about several things at the same time. The complexity of understanding the rationale increases quickly as the text becomes longer. In addition, text data are sequences, which naturally allow dynamics of sentiments throughout the documents. But when prototypes are defined at the document-level, such finer-grained understanding is not possible, and it is difficult for users to relate sentiments to individual sentences.
Second, existing methods generate a large number of prototypes, which is difficult for end-users to comprehend. For example, a ProSeNet model \citep{ming2019interpretable} needs to use hundreds of prototypes to achieve reliable performance. The ProtoAttend \citep{arik2020protoattend} adds sparsity regularization in the model design. But while the number of prototypes involved in each prediction is small, the total number of prototypes the model generates, used by different inputs, is still large, since prototypes are defined on the entire training set. This means users still need to examine $\Omega(|\mathcal{D}|)$ prototypes ($\mathcal{D}$ is the training set) when making predictions.

\begin{figure}[t]
\centering
 \includegraphics[width=\linewidth]{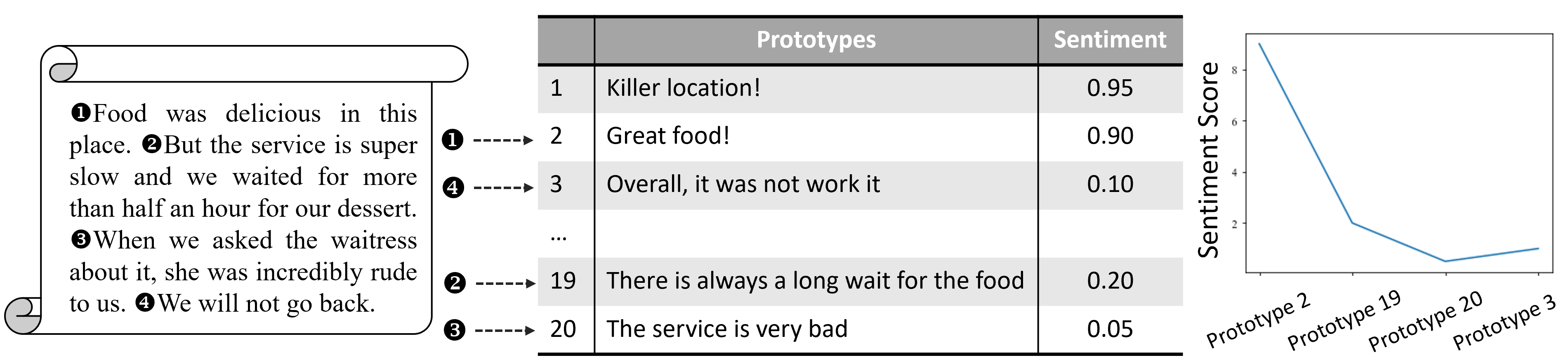}
 \vspace{-3mm}
  \caption{Prototype trajectory-based explanation.}\label{fig:teaser}
\end{figure}
To improve from the two aspects above, we design a new type of prototype-based DNN model, which makes the reasoning process more suitable for text data and uses much fewer prototypes in total.  See Figure 
\ref{fig:teaser} for an example. Each sentence is mapped to only one prototype. Thus, we can relate the sentences to the corresponding sentiments obtained by the model, generating a trajectory of prototypes as well as a trajectory of sentiments. \textcolor{black}{We then use the method proposed in the recent work of \cite{hong2022adaax} to summarize the main trajectory patterns captured by the LSTM part following the prototype layer.} We explain the motivations of the designs below.

Motivated by the nature of text data being sequences, we propose a new concept: \emph{prototype trajectory}, which defines prototypes at the sentence level. The prototype proximity values are then fed into an RNN backbone, which then captures latent patterns across sentences via the trajectory of prototypes.  Prototype trajectories, therefore, illuminate the semantic structure of a text sequence and the logical flow therein and, hence, provides a highly intuitive and useful interpretation of how the model has predicted an output.
In fact, the prototype theory in modern linguistics provides a strong justification for the proposed idea. In the prototype theory, linguists view ``a sentence as the smallest linguistic unit that can be used to perform a complete action'' \citep{alston1964}, analyzing texts with individual sentences as building blocks. Linguists assume that the sentences of a category are distributed along a continuum: at one extreme of this continuum are sentences having a maximal number of common properties, while on the other extreme are sentences that have only one or very few of such properties \citep{panther2008}. Here, the ``ideal'' sentence possessing the maximally shared common properties can be considered as a \textit{prototypical sentence} or a \textit{prototype} of the category. Thus, this paper takes a meaningful first step towards mathematically formalizing the prototype theory in modern linguistics and its analysis methods by incorporating the above view into a computational framework and emulating how linguists analyze a text.

Additionally, to reduce the number of prototypes used for each prediction and the total number of prototypes generated by the model, we introduce two designs to the model. First, each sentence in an input document is mapped to only one prototype, referred to as the \emph{active prototype} for that sentence. This design significantly simplifies the explanations since only $T$ prototypes are used in explaining a prediction,  where $T$ is the number of sentences in the document. Thus, an input document can be represented by a sequence of prototypes.  The idea bears similarity to the ``winner-take-all'' mechanism in competitive learning \citep{rumelhart1985feature,chung1994fuzzy}, where a fundamental module in these neural net models involves taking an input computing its similarities to a collection of prototypes and then selecting only the most similar prototype to ``activate''.  Our experiments show using one active prototype for each sentence performs similarly to using all prototypes, with approximately a 1\% drop in accuracy, while greatly improving understandability.   Second, to reduce the total number of prototypes used by the model, we introduce \emph{prototype pruning} in our proposed model, which prunes away prototypes that are never or rarely mapped to and then retrains the model with the remaining prototypes. Our experiments find that even when the model initializes with 200 prototypes, we end up pruning the majority of them without compromising the predictive performance at all.  {\color{black}For all datasets we used, our model uses about 20 prototypes} while achieving the same predictive performance as using 200 prototypes.

With our design, ProtoryNet permits a fine-grained understanding of sequence data alongside an intuitive explanation of the dynamics of the subsequences, while being simpler to understand than baselines. Since the technical details are hidden in the prototypes, a non-technical user can easily understand the interpretation.

% {\color{black} move this to related work}
% The existing prototype-based methods, however, find the prototypes at the whole-document level, making it difficult to break down the analysis to an individual sentence level, \textit{i.e.}, the connections and flows of individual sentences constituting a paragraph. Moreover, there may not be a suitable prototype when the sequence is too long, as longer sequences have greater degrees of freedom and, thus, harder to find a matching prototype, as evidenced in Section \ref{sec:exp}.

% {\color{black} move this to related work}
% Here, we advocate the idea that the sentence level prototyping (as opposed to the paragraph level prototyping in the previous literature) produce more desirable outcomes, namely better interpretability and higher prediction accuracy. We propose a novel architecture, called \textit{ProtoryNet}, in which we introduce a new concept of the \textit{prototype trajectory}. Given one or more paragraphs, ProtoryNet looks up the nearest prototype for each sentence and computes the proximity. The prototype proximity values are then fed into an RNN backbone, which then captures latent patterns across sentences via trajectory of prototypes. Prototype trajectories, therefore, illuminate the semantic structure of a text sequence and the logical flow therein, and, hence, provides highly intuitive, useful interpretation of how the model has predicted an output, as demonstrated in Figure \ref{fig:teaser}.

% In summary, we make the following contributions ....

The rest of the paper is organized as follows. Section \ref{sec:related} discusses related work, and in particular, compares ProtoryNet with the closest prototype-based DNN for text classification. Section \ref{sec:model} presents the architecture of the ProtoryNet model and Section \ref{sec:train} describes the training procedure. We show detailed experimental results in Section \ref{sec:exp} and human subjects evaluation of the interpretability of ProtoryNet in Section \ref{sec:human} while comparing with another interpretable baseline. Finally, Section \ref{sec:conclusion} concludes the paper and discusses possible future directions. % However, when necessary, technical users, \textit{i.e.}, the ones that are more knowledgeable about RNNs, can still look at the coefficients in RNN, similar to how the attention approaches visualize RNNs, as the proximity vectors feeding the RNN backbone are essentially one-hot encoded (\textit{i.e.}, zero everywhere except the $k$-th position for prototype $k$), making it convenient to trace how coefficients are related to each prototype.

\section{Related Work}\label{sec:related}
% {\color{magenta}{related work is too long}}
% In addition to model-agnostic black-box explainers such as LIME \citep{ribeiro2016should} and SHAP\citep{lundberg2017unified},   
We first review post-hoc explanation methods and attention mechanisms for explaining DNN models, and then we discuss prototype-based DNN in depth and compare it with our proposed model.

\paragraph{Post-hoc Explanations and Attention-Based Methods}
Various post hoc explanation methods have been proposed for explaining DNN models, such as Integrated Gradients \citep{sundararajan2017axiomatic}, DeepLift \citep{shrikumar2017learning}, NeuroX \citep{dalvi2019one}. 
 Specifically, to understand RNN models,  \citet{tsang2018can} proposes a hierarchical explanation for neural networks to capture interactions, and  \citet{jin2019towards} adapts the idea to text classification to quantify the importance of each word and phrase. For sentiment analysis,  \citet{murdoch2018beyond} proposes a contextual decomposition method for analyzing individual predictions made by LSTMs, which identifies words and phrases of contrasting sentiment and how they are combined to yield the LSTM’s final prediction.  
In addition to the external explanation methods, many have considered attention-based approaches interpretable. % the prior efforts to bring interpretability to RNNs can be categorized as \textit{attention-based} and \textit{prototype-based} approaches.  
For example,  \citet{bahdanau2014neural}  implemented an attention mechanism in a decoder, which weighs which part of the source sentence the model needs to pay attention to. 
Similarly, 
% Rockt\"{a}schel \textit{et al}.
~\citet{rocktaschel2015reasoning} analyzed word-to-word attention weights for achieving insights into how a long short-term memory (LSTM) classifier reasons about entailment. %Brown \textit{et al}.~\cite{brown2018recurrent} used the attention mechanism to explore what factors the RNN models pay attention to when predicting anomaly scores. 
% Zhang \textit{et al}.
% ~\cite{zhang2017mdnet} proposed a language model to read and explore discriminative image feature descriptions from reports to learn a direct mapping between lexical components and image pixels via attention. 
Similar strategies can be found in a number of other works (\textit{e.g.} \cite{ismail2019input,choi2016retain}). However, recent research has found attention-based methods controversial, and many works believe they are not explanations \citep{jain2019attention}. In addition, the attention-based approaches are mostly intended for expert users. Many non-technical users in the real world, who lack basic knowledge of how RNNs work (or even neural networks in general), may find them difficult to understand.  

\paragraph{Prototype-based DNN}
% Recently, prototype-based DNN models have been proposed that can achieve close to state-of-the-art performance while being inherently interpretable.  
Prototype-based approaches argue that the intuitiveness of interpretation can be significantly enhanced by visualizing the reasoning process in terms of prototypes. In fact, prototype-based reasoning has a long history as a fundamental interpretability mechanism in traditional models \citep{cupello1988managing, fikes1985role, kim2014bayesian}. One of the first works that introduce prototypical learning into a deep neural network is
~\cite{chen2019looks}, which designed a new neural network architecture for image classification. A prototype layer was added after convolutional layers to compare the convolution responses at different locations with prototypes. From this, users can understand, for example, a bird is classified as a `red-bellied woodpecker' because it has the typical prominent red tint at the belly and the top of its head, as well as the black and white stripes on its wings. The idea was later extended to process video games, using prototypes to explain a player's actions \citep{NEURIPS2022_ae5bf4f3}.

We discuss two prototype-based DNN for text classification. The first model is ProSeNet \citep{ming2019interpretable}, which first uses a sequence encoder to obtain a representation of an input sequence, then uses a prototype layer similar to the one in \cite{chen2019looks} to compare it with a set of prototypes. ProSeNet computes the similarities between an input sequence (usually a short prose) and prototypes and produces the final prediction as a linear combination of the similarities.  Another more recent work is ProtoAttend \citep{arik2020protoattend} which can work with image, text, and tabular data. ProtoAttend utilizes an attention mechanism to select prototypes, and it allows interpretation of the contribution of each prototype via the attention outputs. Similar to ProSeNet, ProtoAttend also relates an input to a linear combination of multiple prototypes. 

\paragraph{Issues We Aim to Solve} Two potential issues might arise in practice for the two models above. First, the prototypes are defined at the document level, therefore when the text is too long, it will be difficult to represent the input with a prototype, and it will be difficult to convince users of their similarity.  The original paper of ProSeNet \citep{ming2019interpretable} validates ProSeNet only on paragraphs shorter than 25 words. However, it is easily fathomable that ProSeNet may fail to assimilate long paragraph data due to large degrees of freedom that complicate the matching of a prototypical example, as validated in our experiments. This may render some practical concerns. For instance, in sentiment classification, even if a paragraph is classified as ``negative,'' it could consist of several twists of sentiments along with sentences (\textit{e.g.}, sarcastic use of positive proses). With an increased length, such kinds of twists can get harder to represent with a prototype, thus making the interpretation difficult and the explanation less credible. This claim is further supported by findings in modern linguistics, which suggest that sentences, instead of paragraphs, should be regarded as the basic elements for text analysis \citep{panther2008}. A second potential issue is the number of prototypes produced, which determines the complexity of the explanations. ProSeNet needs to use $K$ prototypes to explain a prediction, and according to the original paper \citep{ming2019interpretable}, $K$ is at the scale of hundreds. ProtoAttend attends to this issue by including a sparsity regularization in the form of entropy in the training objective. This will make sure there are only a few active prototypes for each prediction. However, the total number of prototypes the model needs to store is at the scale of the size of the training data, which means human users may still need to manually validate and understand all these prototypes when using the model.

ProtoryNet solves the first issue by defining the prototype at the sentence level and solves the second issue by designing specific training objectives and prototype pruning procedures, which will be presented in detail in the next section.

\section{ProtoryNet}\label{sec:model}
We present the architecture of ProtoryNet, describe components in the model and then formulate the learning objective.

\subsection{The ProtoryNet Architecture}
Suppose we work with a data set $\mathcal{D} = \left\{ (\mathbf{X}^{(i)}, \textbf{y}^{(i)}) : i=1,\hdots,N \right\}$ of size $N$, comprised of text sequences $\mathbf{X}^{(i)}$ and the corresponding labels $\textbf{y}^{(i)}$. Here, note that the superscript $(i)$ may be dropped for notational convenience hereinafter, unless necessary. Each instance $\mathbf{X}$ can be understood as a sequence of sentences $\mathbf{x}_t \in \mathbb{R}^V$ at $t$-th position, yielding the representation $\mathbf{X} =
\left(\mathbf{x}_t\right)_{t=1}^{T}$, where $V$ is the size of vocabulary and $T := |\mathbf{X}|$ is the number of sentences in the sequence $\mathbf{X}$. 
$\textbf{y} \in \mathbb{R}^{C}$ is a multi-hot encoded vector representing the class labels associated with the sequence $\mathbf{X}$, \textit{i.e.}, the $c$-th element $y_c$ of $\mathbf{y}$ equals 1 if the label $c$ is associated with $\mathbf{X}$ or 0 otherwise. $C$ is the total number of classes.

\begin{figure*}[t]
    \centering
    \includegraphics[ width=1.03\textwidth]{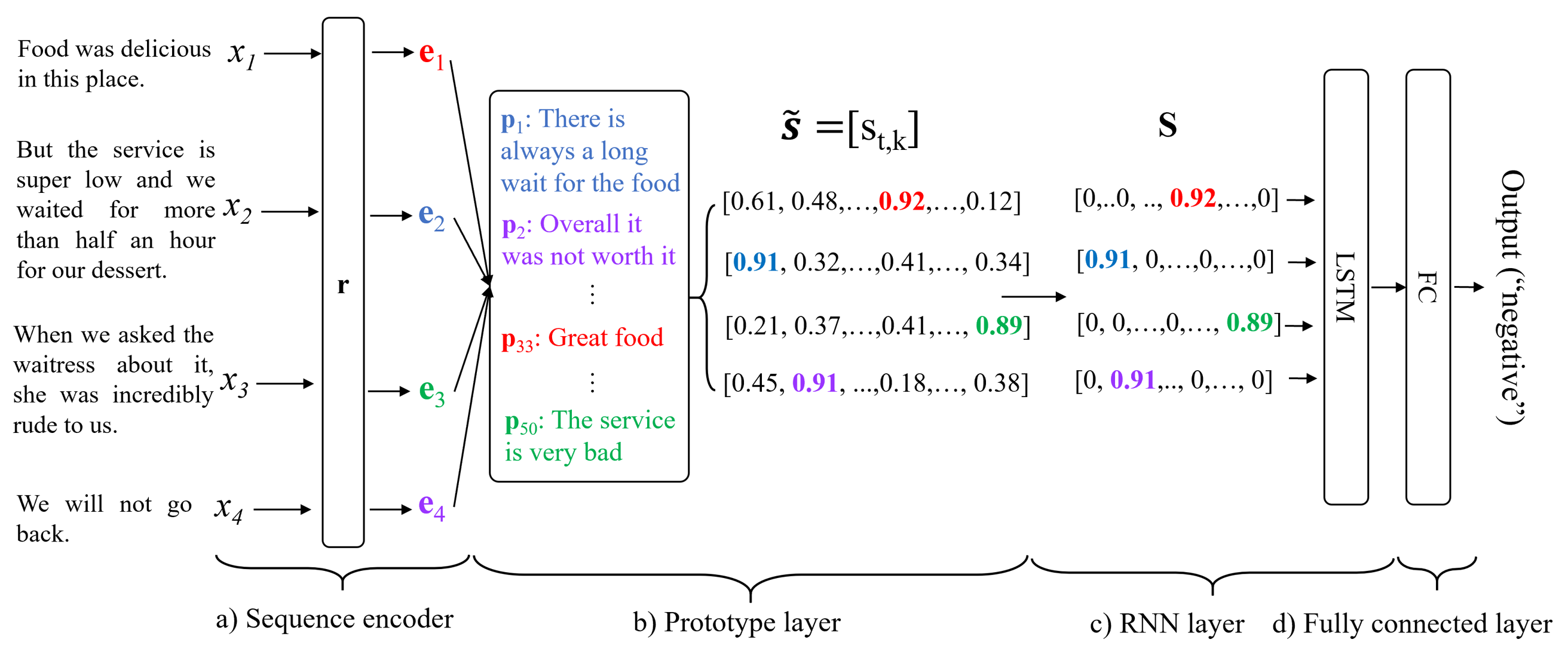}
    
    \caption{The architecture of ProtoryNet.}
    \label{fig:architecture}
\end{figure*}

ProtoryNet interfaces with text data via a sentence encoder (Figure~\ref{fig:architecture}a) modeled as a mapping ${r}: \mathbb{R}^V \rightarrow \mathbb{R}^J$, where $J$ is the dimension of sentence encoding specified by the user. That is, the encoder takes each sentence $\mathbf{x}_t \in \mathbf{X}$ and produces a sentence embedding:
\begin{equation}
 \textbf{Sequence Encoder Layer }: \quad   \mathbf{e}_t = {r}(\mathbf{x}_t).
\end{equation}
The development of the encoder $r(\cdot)$ is beyond the scope of this paper and, hence, we employ a state-of-the-art Transformer encoder, Google Universal Encoder  \citep{cer2018universal}, where $J=512$ by default. 
The encoder layer may or may not be fine-tuned, which will have an impact on the predictive performance. For now, we defer the discussion to Section \ref{sec:accuracy}.

The sentence embeddings $\mathbf{e}_t$ are then fed into the \textit{prototype layer} (Figure~\ref{fig:architecture}b), in which a set of trainable prototypes $\mathcal{P}=\left\{ \mathbf{p}_k \in \mathbb{R}^J : k=1,\hdots,K\right\}$ are compared with $\mathbf{e}_t$, where $K := \left|\mathcal{P}\right|$ is the number of prototypes specified by the user, and each prototype vector $\mathbf{p}_k$ has a dimension of $J$. Note that prototypes $\mathcal{P}$ are trainable parts of the model. 
Then, given a distance metric $d: \mathbb{R}^J \rightarrow \mathbb{R}^+$, the proximity $s_{t,k}$ of the sentence embedding $\mathbf{e}_t$ to a given prototype $\mathbf{p}_k$ is measured as 
\begin{equation}
\textbf{Prototype Layer}:    s_{t,k} :=  \exp \left(-\frac{ d(\mathbf{e}_t, \mathbf{p}_k)}{\psi^2} \right),
\end{equation} 
where $\psi \in \mathbb{R}$ is a user-specified constant which we set it to be $\psi^2 = 10$. Between two popular choices for the distance metric $d(\cdot)$, namely the cosine distance and the Euclidean distance, we find that there is no significant difference between the two. Hence, we use the Euclidean distance in our experiments for convenience.

Note that the intermediate throughput of the prototype layer is the similarity matrix $\tilde{\mathbf{S}}=\left[ s_{t,k} \right]$ of the size $T \times K$, associating the $t$-th sentence with the $k$-th prototype. The rows of the similarity matrix $\tilde{\mathbf{S}}$ then constitute the input to the LSTM backbone at time step $t$ (Figure~\ref{fig:architecture}c), which then finally produces an output prediction. However, doing so means each sentence is associated with all $K$ prototypes. With the total number of sentences being $T$, the explanation will then involve $T\cdot K$ prototypes. To ensure better interpretability, we would like to generate easy explanations where each sentence is mapped to only one prototype instead of $K$ prototypes. And we want it to be the most similar prototype to make the explanation more convincing. This means we need to set each row in $\tilde{\mathbf{S}}$ to zero except for the position where $s_{t,k}$ is the maximum. That is, each row of the transformed similarity matrix $\mathbf{S}$ would be of the same topology as the one-hot encoded vector, whose elements equal $s_{t,k^*}$ at $k^*:=\argmax_k s_{t,k}$ and 0 otherwise.  For future reference, we denote the most similar prototype for a given sentence the \textbf{active prototype}. In this case, prototype $k^*$ is the active prototype for the $t-$th sentence. %Then LSTM model only considers the active prototype for each sentence when making a prediction. %We call it \textit{sparsity transformation}. 

The sparsity transformation, unfortunately, is not differentiable and may lead to an unexpected training behavior during auto-differentiation in deep learning packages. We get around this issue by the following approximation technique. Suppose the similarity matrix $\tilde{\mathbf{S}}$ = 
$[\tilde{\mathbf{s}}_1, \hdots, \tilde{\mathbf{s}}_T]^\top$, 
where $\tilde{\mathbf{s}}_t \in \mathbb{R}^K$ is a row vector whose elements indicate how similar the $t$-th sentence is to each of the prototypes. If we let $\text{Softmax}(\cdot)$ to denote the softmax function, then for some large constant $\gamma$,
\begin{equation}
\mathbf{\Gamma} = [\text{Softmax}(\gamma \cdot \tilde{\mathbf{s}}_1),..., \text{Softmax}(\gamma \cdot \tilde{\mathbf{s}}_T)],
\end{equation}
which approximates the selection matrix whose element equals to 1 at the position corresponding to where $\mathbf{s}_t$ is the maximum for each column $t$ and 0 elsewhere. Here, we find $\gamma \ge 1e^6$ gives a reasonable approximation empirically. With the selection matrix, the sparsity transformation can be approximated as follows without explicitly computing the maximum:
\begin{equation}\label{eqn:S}
\textbf{Sparsity Transformation}: \quad
\mathbf{S} 
\approx
\mathbf{\Gamma} \odot \tilde{\mathbf{S}},
\end{equation}
where $\odot$ is the Hadamard product. Note that the softmax function is differentiable and, thus, is $\mathbf{S}$.

% Each row of the transformed similarity matrix $\mathbf{S}$ would be of the same topology as the one-hot encoded vector, whose elements equal $s_{t,k^*}$ at $k^*:=\argmax_k s_{t,k}$ and 0 otherwise.

The sparsity transformation of $\tilde{\mathbf{S}}$ to $\mathbf{S}$ enhances the interpretability of the architecture, by enforcing each sentence to be matched with the most similar prototype and, thus, disentangling the information. This is accomplished only at a small cost of accuracy, as observed from an ablation study in Section \ref{sec:ablation}. %Later, we contend this idea by comparing the performance of ProtoryNet with versus without the transformation in the Appendix to test if there is any loss in accuracy by forcing the sparsity.
 Since an input text now can be regarded as a sequence of prototypes, one can think of the matrix $\tilde{\mathbf{S}}$ as a type \textbf{prototype encoding} and matrix $\mathbf{S}$ is a sparse prototype encoding. Unlike other sequence encoders (e.g., using embedding techniques) that yield feature vectors that are not sensible to humans, here the prototype encoding returns features (i.e., prototype) that are meaningful and easily understandable. 

Next, each row of $\mathbf{S}$ is fed into an LSTM model, followed by a few fully connected layers, denoted as,
\begin{align}
\textbf{RNN Layer}:&\quad    z = \gamma(\mathbf{S}),\\
\textbf{Fully Connected Layer}:&\quad \hat{y} = \phi(z),
\end{align}
where $\gamma(\cdot)$ represents the LSTM layer and $\phi(\cdot)$ represents a fully connected layer transformation. %The LSTM model only considers the active prototype for each sentence when making a prediction. 

\paragraph{Motivating Example}  We present an example to further demonstrate the model. The text data in Figure~\ref{fig:architecture} exemplifies the use of ProtoryNet for sentiment analysis (text classification). In this example, the task is to predict whether the review of a restaurant is positive or not. The input text data $\mathbf{X}$ is comprised of $T = 4$ sentences, in this particular case, and the label $\textbf{y}$ is the binary sentiment label of the review, either ``positive'' ($[1,0]$) or ``negative'' ($[0,1]$). ProtoryNet converts the text data into sentence embeddings, each of which is then matched with the closest prototype. Observe, in the figure, that the prototypes that ProtoryNet produced are, indeed, morphosyntactically equivalent to the corresponding input sentences, well-exemplifying them semantically. The one-hot-like similarity vectors between the sentences and the prototypes are then fed into the LSTM backbone, which captures the patterns and trends in the trajectory of prototypes and, finally, predicts the final sentiment label, which, in this case, is ``negative.''

\subsection{Objective Functions}
The training objectives of ProtoryNet entail four different terms aiming to achieve both prediction accuracy and interpretability. Below are the details of their definitions.

\paragraph{Accuracy} The \textit{accuracy loss} is defined as the mean squared loss between the predicted value and the ground truth label, promoting the model to make accurate predictions:
\begin{equation}
\mathcal{L}_\text{acc} (\mathcal{D}) := \frac{1}{N} \sum_{i=1}^{N} {\left\|\textbf{y}^{(i)} - \hat{\textbf{y}}^{(i)} \right\|^2}.
\end{equation}

\paragraph{Diversity} To ensure diverse and non-overlapping prototypes, % When the number of prototypes $K$ is large, prototypes will become closer to each other. exhibits a tendency to cluster closely to each other. ({\color{red}This claim needs to be validated in the result section.}){\color{black}That's what we observed long before when we just started our experiments. Do we need to add something in the "experiment" section?} To avoid this, 
we define the \textit{diversity loss} term added to enforce the minimum mutual distance $\delta$ among the prototypes:
% \begin{equation}
% C_2 = \left\{ 1 + e^{a\left( \min_{k_1,k_2 \in [1,K]}d(\textbf{p}_{k_1},\textbf{p}_{k_2}) - d_{min} \right)}\right\} ^{-1}
% \end{equation}
\begin{align}
d_\text{min} &:= \min_{k_1,k_2}d(\mathbf{p}_{k_1},\mathbf{p}_{k_2}),
\\
\mathcal{L}_\text{div} (\mathcal{D}) &:= \sigma\left( \eta(\delta - d_\text{min}) \right),
\end{align}
where $d(\cdot)$ is the Euclidean distance, $\sigma(\cdot)$ is the sigmoid function and $\eta$ is a smoothing constant, which we set $\eta=1$ empirically. The constant $\delta \in \mathbb{R}^+_*$ is a positive real number defined by the user, to enforce the minimum separation among prototypes. Hence, when the distances among the prototypes do not meet the minimum separation requirement \textit{i.e.}, $d_\text{min} < \delta$, the $\eta(\delta - d_\text{min})$ term will have some positive value, making the diversity loss term $\mathcal{L}_\text{div}$ active; on the other hand, when the minimum separation requirement is met and thus, $d_\text{min} > \delta$, then the sigmoid function will pull the loss term to zero. Note that a smaller $\eta$ will make such a transition by the sigmoid function smoother.
% \textcolor{red}{double check d and $\delta$}

\paragraph{Prototypicality}
With only the accuracy and the diversity terms alone, it is observable a prominent tendency of prototypes diverging away from the sentences during training. Such a behavior introduces overfitting, in which prototypes become less generalizable, as the prototypes lose their representativity of a category. In addition, it also hurts the prototypicality of the prototypes since the prototypes are too far away from the sentences to properly represent the sentences.  Hence, we introduce the \textit{prototypicality} loss, which promotes each sentence in the database to have a representative prototype close to it, i.e., we encourage the distance between a sentence and its active prototype to be small:
\begin{equation}\label{eqn:prototypicality}
\mathcal{L}_\text{proto} = \frac{1}{M}\sum_{\textbf{X} \, \in \mathcal{D}}^{} \sum_{\mathbf{x}_t \in \mathbf{X}} \, \min_{k}   d(\textbf{r}(\mathbf{x}_t),\textbf{p}_k),
\end{equation}
where $M$ is the total number of sentences in the data set.

% \paragraph{Coverage}
% It is likely only a few prototypes are used instead of all prototypes. While the prototypicality term makes sure each sentence is close enough to at least one prototype, it is possible that they are all close to a subset of prototypes, leaving the rest of the prototypes unused. Therefore, we would want to t
% \begin{equation}
% \mathcal{L}_\text{cov} = \frac{1}{M}\sum_{\textbf{X} \, \in \mathcal{D}}^{} \sum_{\mathbf{x}_t \in \mathbf{X}} \, \min_{k}   
% \end{equation}
% However, if we only miminize $C_3$, then the subsequences will pick only one prototype and ignore others. To achieve the interpretability for all prototypes, we also minimize another evidence regularization loss function:

% \begin{equation}
% C_4 = \sum_{\textbf{\textit{X}} \, \textrm{in} \, \CMcal{D}}^{} \sum_{k=1}^{K} \, \min_{t = 1}^{T}   d(\textbf{e}_t,\textbf{p}_k)
% \end{equation}

% These $C_3$ and $C_4$ will be jointly minimized in our training process. 
% In our experiments, we also divide $C_3$ and $C_4$ by $J$ to reduce the scale.

\paragraph{Final loss}
The final loss function combines the above loss terms:
\begin{equation}
\mathcal{L} = \mathcal{L}_\text{acc} + \alpha \mathcal{L}_\text{div} + \beta \mathcal{L}_\text{proto}.
\end{equation}
Empirically, coefficient values of $\alpha={0.1}$ and $\beta=1e^{-4}$ are used by default in this paper except in the sensitivity analysis.

\paragraph{Remarks on Prototype Interpretability} The diversity and prototypicality terms are designed for improving interpretability. Here, to achieve good explanations, prototypes need to be different from each other to avoid redundancy, thus the diversity term. In addition, each input sentence needs to be mapped to a prototype that is similar enough to make the explanation convincing, thus the prototypicality term. These two loss terms can be considered regularization terms to serve interpretability purposes. Similar loss terms have been introduced in other prototype-based DNN models \citep{ming2019interpretable,chen2019looks}. We will later show in experiments that these two terms do not hurt the predictive performance. This can be explained by the recent research on ``Rashomon Set'' \citep{semenova2019study, rudin2019stop}, that there exist many models with very similar performance, so one can add customized constraints to the model to achieve additional benefits, such as interpretability.

\section{Training a ProtoryNet}\label{sec:train}
For the training of ProtoryNet, the adaptive moment estimation (ADAM) optimizer \citep{kingma2014adam} was employed. The learning rate was set to be $1e^{-4}$ and the exponential decay rates for the first and the second-moment estimates were 0.9 and 0.999, respectively. Below are further details used for generating the results in this paper.

\subsection{Prototype Initialization}
The training of ProtoryNet can benefit from the initialization method described below. We first embed all sentences separately in the training data set. Then, in the embedding space, all sentences in the data set are clustered using the $k$-medoids clustering algorithm to categorize sentences by their semantic meaning. The medoids obtained from the $k$-medoids algorithm can be considered as representative examples of each cluster and, hence, plausible candidates for prototypes. Thus, for the training of ProtoryNet, we use these medoids to initialize prototypes, which in turn accelerates the convergence.

\subsection{Prototype Projection}
It should be noted that the numerical solutions for the prototypes are found in the embedding space. These numerical solutions are not automatically intelligible to human users and need to be deciphered. To this end, we project the prototypes to the closest sentence from the training data in the embedding space every 10 epochs during the training process, similar to the technique proposed in  \cite{ming2019interpretable,chen2019looks}:
% {\color{black} I cited our 2 referenced papers: ProSeNet and "it looks like that..." }
\begin{equation}\label{eqn:projection}
\mathbf{s}_k = \argmin_{\mathbf{x}_t \in X^{(i)}, \forall X^{(i)} \in \mathcal{D}} d(\mathbf{r}(\mathbf{x}_t),\mathbf{p}_k), \qquad k \in [1,K].
\end{equation}

\subsection{Prototype Pruning} 
In our analysis and experiments, we find that prototypes have significantly different probabilities to be selected (mapped to as the \emph{active prototype}).  While the prototypicality term makes sure each sentence is close enough to at least one prototype, we observe that sentences are usually close to a small subset of prototypes, leaving the rest  rarely or even never ``activated'' in inference.

\begin{figure*}[h]
\centering
 \includegraphics[width=0.76\linewidth]{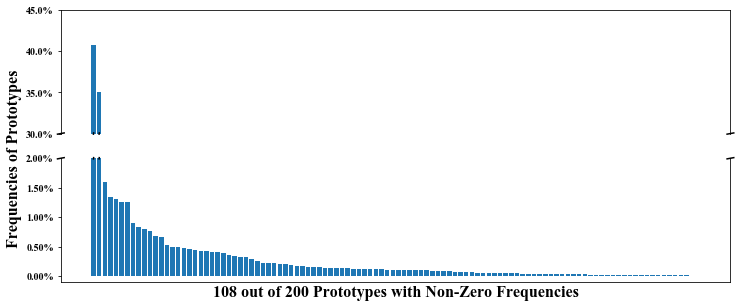}
 \caption{The frequencies of prototypes from a ProtoryNet model trained on Amazon dataset with K = 200}\label{fig:frequencies}
\end{figure*}

For demonstration, we show an example from the experiment section later in the paper. This model is trained on the Amazon dataset, and the original $K$ is set to 200. We compute the frequencies of prototypes being active for the model trained on the Amazon dataset. Out of 200 prototypes, 92 prototypes have never been mapped to by any sentences in the validation set, which means that these prototypes can already be pruned away without affecting performance. Then, we plot the frequencies of the remaining 108 prototypes in Figure \ref{fig:frequencies}. The prototypes are ranked in descending order of frequencies of being active, i.e., the left-most prototype has the highest frequency: more than 40\% sentences are mapped to this prototype, while the right-most prototype has the lowest frequency of less than 0.01\%. We observe that the frequencies decay rapidly, indicating that only the top-ranked prototypes are heavily used by the model. %We observe similar trend for  other dataests too. There are often a large prototypes unused or rarely used. This is an encouraging observation that  justifies the idea of prototype pruning, that in addition to its obvious benefit of improving the model interpretability, prototype pruning seems to reduce the redundancy in the model.

We observe similar patterns in other datasets where only a subset of prototypes are used, and the remaining are never activated or rarely activated. This is an encouraging observation that justifies the idea of prototype pruning, that in addition to its obvious benefit of improving the model interpretability, prototype pruning seems to reduce the redundancy in the model without hurting the performance.
We hypothesize that this is due to the diversity term in the objective, which keeps prototypes distant from each other. Then, when only a few prototypes are sufficient for covering the data space, redundant prototypes are pushed away from all sentences since they need to remain $\delta$ away from other prototypes. Because of this, we propose to do prototype pruning, which is to remove these redundant prototypes after the training is complete, based on their frequencies of being active, evaluated on a validation set. If the frequency is smaller than a threshold $\theta$, then the prototypes are removed. The steps are described in lines 11 and 12 in Algorithm 1. Let $\mathcal{K}$ represent the indices of remaining prototypes. $\mathcal{K}\subset \{1,2,\cdots, K\}$ and $|\mathcal{K}| = \hat{K}$. The remaining prototype vectors are $\{\mathbf{p}_k\}_{k\in \mathcal{K}}$. In practice, the threshold $\theta$ can be tuned via a validation set.% and the corresponding prototypical sentences are represented by $\{\mathbf{s}_k\}_{k\in \mathcal{K}}$ via prototype project from equation (\ref{eqn:projection}).

When implementing the prototype pruning, we build a new ProtoryNet, denoted as $\tilde{f}$. $\tilde{f}$ consists of the same sentence encoder layer $r(\cdot)$ and the $\hat{K}$ prototypes $\{\mathbf{p}_k\}_{k\in \mathcal{K}}$ selected above. We freeze $r(\cdot)$ and $\{\mathbf{p}_k\}_{k\in \mathcal{K}}$ and allow the rest of the layers in $\tilde{f}$ to be trained, i.e., the LSTM layer, denoted as $\tilde{\gamma}(\cdot)$, and fully connected layer, denoted as $\tilde{\phi}$. The steps are described from line 14 to 19 in Algorithm 1. %Note that the goal of prototype pruning is to improve interpretability by keeping a smaller number of prototypes instead of $K$.  %Since the redundant prototypes are not used or rarely used by the model in the first place, this pruning step does not hurt the predictive performance, as will be shown in the experiments.

\paragraph{Sentiment Scores for Prototypes } 
Once the training is done, ProtoryNet returns a set of $\hat{K}$ prototypes and $\hat{K} < K$. We then feed the prototypes back into the trained ProtoryNet one at a time. The outputs from the model are the corresponding sentiment scores of each prototype. These sentiment scores will later be used to provide quantitative visualizations of how the tones and sentiments change within text data.

We summarize the training procedure in Algorithm \ref{alg:train} \footnote{Code can be found at \url{https://github.com/dathong/ProtoryNet}}.

\begin{algorithm}[ht]
 \caption{Training Procedure for ProtoryNet}\label{alg:train}
 \begin{algorithmic}[1]
\STATE \textbf{Input}: $K$, $\mathcal{D}_\text{train}, \mathcal{D}_\text{val}, \alpha, \beta, \delta, \theta$, FineTuning\\
\STATE \textbf{Initialization}: 
Build a ProtoryNet $f = \{r, \{\mathbf{p}_1, \cdots, \mathbf{p}_K\}, \text{LSTM-layer}, \phi\}$ and set $r, \mathbf{p}_1, \cdots, \mathbf{p}_K$, LSTM-layer, and $\phi$ to trainable
\STATE \textcolor{gray}{\# -------------------- this block trains the model to obtain $K$ prototypes --------------------}
     \IF{FineTuning = \emph{FALSE}}  
      \STATE$r(\cdot)$ $\leftarrow$ non-trainable 
    \ENDIF
\FOR{$j\gets 0 $ to $n_\text{epoch}$}
\STATE train $f$ with ADAM  
\ENDFOR
% \STATE \textcolor{gray}{\#  ---------------------  this block fine tunes the model after projection  ---------------------}
% \STATE $\mathbf{p}_k = r(\mathbf{s}_k), k = 1, \cdots, K$ \COMMENT{\textcolor{gray}{freeze $s_k$ but set the encoder $r$ according to FineTuning. }}
%  \COMMENT{\textcolor{gray}{RNN and the fully connected layer remain trainable}}
% \FOR{$j\gets 0 $ to $n_\text{epoch}^\prime$}
% \STATE train $f(\cdot)$ with ADAM
% \ENDFOR
\STATE \textcolor{gray}{\# ------------------------------------------  prototype pruning  ------------------------------------------}
\STATE Compute the frequencies of active prototypes using $\mathcal{D}_\text{val}$
\STATE
Select $\{\mathbf{p}_k\}_{k\in\mathcal{K}}$ whose frequencies are larger than $\theta$. 
\STATE \textcolor{gray}{\# --------------------------  retrain the model with $\hat{K}$ prototypes fixed  --------------------------}
\STATE Build a new ProtoryNet $\tilde{f} = \{r, \{\mathbf{p}_k\}_{k\in\mathcal{K}}, \tilde{\gamma}, \tilde{\phi}\}$, where the $r(\cdot)$ and $\{\mathbf{p}_k\}_{k\in\mathcal{K}}$ are identical to those in $f$ 
\STATE  $r(\cdot)$, $\{\mathbf{p}_k\}_{k \in \mathcal{K}}$ $\rightarrow$ non-trainable
\STATE $\tilde{\gamma}, \tilde{\phi} \rightarrow$ trainable
\FOR{$j\gets 0 $ to $n_\text{epoch}$}
\STATE train $\tilde{f}(\cdot)$ using ADAM
\ENDFOR
\STATE Evaluate the sentiment scores for each prototypes $\mathbf{p}_k = r(\mathbf{s}_k), |\mathcal{K}| =  \hat{K}$.
\STATE $\{\mathbf{s}_k\}_{k=1}^K \leftarrow$ Prototype projection of $\{\mathbf{p}_k\}_{k=1}^K$ using Formula (\ref{eqn:projection})
\STATE \textbf{Return}: $\hat{f}(\cdot),\{\mathbf{s}_k\}_{k=1}^K$
\end{algorithmic}
\end{algorithm}

\section{Experiments}\label{sec:exp}
In this section, we evaluate ProtoryNet on six data sets (a detailed description and data preparation for the datasets are included in Section \ref{sec:datasets} in the Appendix).  Our method is compared against a vanilla LSTM method, an accurate black-box model, DistilBERT \cite{sanh2019distilbert}, and a state-of-the-art prototype-based interpretable model, ProSeNet \cite{ming2019interpretable}. We also compare our method with a non-neural bag-of-words baseline, which can provide explanations at a word level. See a description of the  model setup in Section \ref{sec:models} in the Appendix. 

Our goal is to investigate whether ProtoryNet is comparable to other interpretable baselines and how much accuracy it may lose compared to the black-box model. Then, we analyze the effect of prototype pruning on the prediction performance. In addition, we provide a detailed example of ProtoryNet using one of the datasets and demonstrate the complexity of sentiment trajectories. Finally, we study the effect of various hyper-parameters in the model.

\subsection{Prediction Accuracy}
\label{sec:accuracy}
We foremost demonstrate that our inherently-interpretable model design does not cause significant degradation in performance while beating other interpretable baselines

 We implement two types of ProtoryNet in the experiments, a \textbf{fine-tuned} version where the sentence encoder continues to be trained on the target dataset with the rest of ProtoryNet and a \textbf{non-fine-tuned} version where the universal sentence encoder is used as a service but not updated during training. The non-fine-tuned ProtoryNet needs to train significantly fewer coefficients, about 0.03\% of the fine-tuned models, thus consuming less energy and computing resources. The goal is to explore 1) the best performance ProtoryNet can achieve, with the help of the state-of-the-art sentence encoder, and 2) the more economical solution for use in resource-constrained scenarios.

Here we keep the parameters same for all datasets:  $K = 200, \alpha = 0.01, \beta = 1e^{-4}, \delta = 1, \eta = 1$. 
Our intention is to show that the method is robust enough to solve different text classification problems with varying complexity using one single model architecture and hyperparameters. This way, the ProtoryNet will be practically accountable and easier to use in practice since it does not necessarily need exhaustive tuning of hyper-parameters. We will later explain in Section \ref{sec:pruning} why ProtoryNet is insensitive to $K$ and investigate its sensitivity  to $\alpha$ and $\beta$ in Section \ref{sec:ablation}. In addition, we also do not do prototype pruning in this part of the experiment and we will investigate it in detail in Section \ref{sec:pruning}.

Reported in Table~\ref{tbl:accuracy} are performance on the six data sets. First, we acknowledge the performance gap compared to the black-box models. As expected, the black box model (DistilBERT) has the best performance in all data sets used. But still, both versions of ProtoryNet reduce the gap. They both outperform the two interpretable baselines and Vanilla LSTM, and if we allow fine-tuning, ProtoryNet becomes even better, with a more significant increase compared to the baselines and only 1.9\% away on average from the black-box  DistilBERT. 

\begin{table*}[t]
\caption{Performance of ProtoryNet in comparison with other benchmark models.}
\label{tbl:accuracy}
\centering
\small
%  \begin{tabular}{@{\hskip 10pt}l@{\hskip10pt} |@{\hskip10pt}c @{\hskip10pt}c |@{\hskip10pt} c@{\hskip10pt} c@{\hskip10pt}  c@{\hskip10pt} c@{\hskip10pt} c}
%     \toprule 
%     Data set & DistilBERT  & \thead{\textbf{ProtoryNet} \\ \textbf{(fine-tuned)}}  & Vanilla LSTM & ProSeNet  & \textbf{ProtoryNet}  & Bag-of-words\\
%     \midrule
%     IMDB & 0.931  & 0.914 & 0.871  & 0.863  &  0.871 & \textbf{0.877} \\
%     Amazon Reviews & 0.940  & 0.918 & 0.884  & 0.875  & 0.890 & 0.830\\
%     Yelp Reviews & 0.967  & 0.954 & \textbf{0.952}  & 0.948   & 0.925 & 0.908 \\
%     Rotten Tomatoes & 0.903  & 0.881  & \textbf{0.877} & 0.869 & 0.771 & 0.785 \\
%     Hotel Reviews & 0.976  & 0.961 & 0.949 & 0.930 & \textbf{0.949} & 0.905\\
%     \bottomrule
%   \end{tabular}
\begin{tabular}{@{\hskip 4pt}l@{\hskip4pt} |@{\hskip4pt}c @{\hskip4pt} |c @{\hskip4pt} c@{\hskip4pt} c@{\hskip4pt}  c@{\hskip4pt} c@{\hskip4pt} c}
    \toprule 
    Data set & DistilBERT  & \thead{\textbf{ProtoryNet} \\ (Fine-tuned)}  & \thead{\textbf{ProtoryNet} \\ (Not fine-tuned)} & ProSeNet  & \thead{Vanilla\\ LSTM} & Bag-of-words\\
    \midrule
    IMDB & 0.931  & 0.914 & 0.871  & 0.863  &  0.871 & 0.877 \\
    Amazon Reviews & 0.940  & 0.918 & 0.890  & 0.875  & 0.884 & 0.830\\
    Yelp Reviews & 0.967  & 0.962 & 0.941  & 0.932   & 0.952 & 0.908 \\
    Rotten Tomatoes & 0.903  & 0.881  & 0.771 & 0.869 & 0.877 & 0.785 \\
    Hotel Reviews & 0.976  & 0.961 & 0.949 & 0.930 & 0.949 & 0.905\\
    Steam Reviews & 0.955 & 0.924 & 0.876 & 0.834 & 0.864 & 0.844 \\
    \bottomrule
  \end{tabular}
  \label{table:acc1}
\end{table*}
\paragraph{Fine-Tuning vs Non-Fine-Tuning} To choose between the fine-tuned and non-fine-tuned ProtoryNet in practice, users need to trade-off between the time and computing resource consumption and the predictive performance. There are more than 256 million parameters in the sentence encoder and only 68 thousand parameters in the rest of ProtoryNet (when $K = 200$), which means that the non-fine-tuned ProtoryNet can be trained only with less than 0.03\% of the parameters compared to the fine-tuned version.  {In addition, training a fine-tuned ProtoryNet takes much longer in time (approximately three times longer on the same Google Colab notebook with a GPU accelerator) than a non-fine-tuned ProtoryNet.}  In summary, the non-fine-tuned ProtoryNet is much smaller and more {energy efficient}, while still beating the interpretable baselines. Answering the increasing call for Green-AI \cite{schwartz2019green}, non-fine-tuned ProtoryNet will be better than the fine-tuned ProtoryNet when smaller, and lighter models are preferred.

\paragraph{Comparing Short and Long Reviews } Between ProSeNet and ProtoryNet, ProtoryNet outperformed ProSeNet for all six cases overall. In particular, the performance difference was clearer when long text data were analyzed. Since the fine-tuned ProtoryNet significantly outperforms ProSeNet, here we only compare ProSeNet with the weaker version of ProtoryNet, the non-fine-tuned models. In Table~\ref{table:acc2}, we split each data set into \textit{short} and \textit{long} samples---texts that were less than 25 words were classified as short samples, following the criterium used in the ProSeNet paper \citep{ming2019interpretable}. As shown in the table, ProSeNet was on par or better than ProtoryNet on short texts, while ProtoryNet was better than ProSeNet when long paragraphs were concerned. In fact, this is an advantage of ProtoryNet since long texts (more than 25 words) are more prevalent than short texts in most real-world datasets, as evidenced in Table \ref{table:acc2}.  This also explains why the non-fine-tuned ProtoryNet performs worse than ProSeNet on the Rotten Tomatoes dataset since more than 65\% of the reviews are short reviews with less than 25 words.
\begin{table}[h]
  \caption{Comparison between ProSeNet and ProtoryNet (non-fine-tuned) on text data of different lengths. } 
  \label{table:acc2}
  \centering
  \begin{tabular}{l | c |c c|  c c}
    \toprule 
    \multirow{2}{*}{\textbf{Data set}} &
    \multirow{2}{*}{\textbf{\% of short reviews} }
    % & \multirow{2}{*}{\% of short samples
    %} 
    &\multicolumn{2}{c|}{\textbf{ProSeNet}} &  \multicolumn{2}{c}{\textbf{ProtoryNet}}\\
    \cline{3-6} &   & Short & Long & Short & Long\\
    \midrule
    IMDB & 0.17
    % \footnote[1]{The ratio of short to total reviews in IMDB is very imbalanced, i.e., 22/25000.} 
    & 0.868  &  0.863  &  0.868 & 0.871 \\
    Amazon Reviews& 6.02 & 0.908  &  0.873   &  0.843 & 0.893\\
    Yelp Reviews& 8.85 & 0.943  & 0.931 &  0.863 & 0.949\\
    Rotten Tomatoes & 65.52& 0.875  & 0.859   & 0.751 & 0.809\\
    Hotel Reviews & 2.11& 1.000  & 0.928 &  1.000 & 0.949 \\
    Steam Reviews& 23.75 & 0.791 & 0.848  & 0.860 & 0.881 \\
    \bottomrule
  \end{tabular} 
\end{table}

\textcolor{black}{\paragraph{Discussion} 
We can compare the performance of interpretable models based on the results from Table \ref{table:acc1} and Table \ref {table:acc2}.
We can see between ProtoryNet and ProSeNet, ProtoryNet outperforms ProSeNet in long text documents. This is because ProSeNet's prototypes are at the document level, so when the review is long (more than 25 words), there is more likely a loss of information when defining prototypes to represent the entire review. ProtoryNet mitigates this problem by splitting a document into sentences, making processing long documents viable. The superior performance of ProtoryNet on long reviews leads to a better performance on average, since most reviews are longer than 25 words in a review dataset, as shown in Table 2.
Compared to other baselines, ProtoryNet (not fine-tuned) performs similarly to Vanilla LSTM and better than bag-of-words because bag-of-words do not consider the sequential nature of text data.}

\subsection{Prototype Pruning}\label{sec:pruning}
In previous experiments, we set K to a fixed number ($K = 200$) for all datasets to show that our method is robust enough to solve different text classification problems using the same parameters. In this section, we apply prototype pruning after a model is trained for the purpose of improving interpretability, since fewer prototypes there are, the easier it is for human users to understand the model and interpret the predictions.

% First, we investigate the utilization of the 200 prototypes from models in Section \ref{sec:accuracy}, i.e., how often each prototype is actually mapped to as the active prototype by the sentences? We compute the frequencies of prototypes being active for the model trained  on the Amazon dataset as an example. Out of 200 prototypes, only 108 have been ever mapped to by any sentences in the validation set. We plot the frequencies of the 108 prototypes in Figure \ref{fig:frequencies}. The prototypes are ranked in a descending order of frequencies,i.e., the  left most prototype has the highest frequency: more than 40\% sentences are mapped to this prototype. We observe that the frequencies decay rapidly, indicating that only the top ranked prototypes are heavily used by the model. We observe similar trend for  other dataests too. There are often a large prototypes unused or rarely used. This is an encouraging observation that  justifies the idea of prototype pruning, that in addition to its obvious benefit of improving the model interpretability, prototype pruning seems to reduce the redundancy in the model without possibly hurting the performance.
% \begin{figure*}[h]
% \centering
%  \includegraphics[width=0.9\linewidth]{figures/frequencies.png}
%  \caption{The frequencies of prototypes from a ProtoryNet model trained on Amazon datsaet with K = 200}\label{fig:frequencies}
% \end{figure*}

\begin{table}[t]
\centering
\begin{minipage}{.5\textwidth}\vspace{-4mm}
  \centering
    \includegraphics[width=0.95\linewidth]{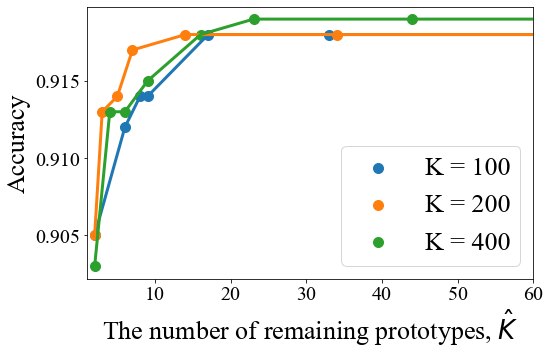}
    \vspace{-4mm}
    \captionof{figure}{The effect of prototype pruning}\label{fig:amazon_pruning}
\end{minipage}%
\quad
\begin{minipage}{.46\textwidth}
\centering
  \begin{tabular}{l| c }
    \toprule 
   \textbf{ Data set}  & \textbf{$\hat{K}$}\\
    \midrule
    IMDB  & 8  \\
    Amazon Reviews & 14 \\
    Yelp Reviews & 15  \\
    Rotten Tomatoes & 23 \\
    Hotel Reviews &  7\\
    Steam Reviews &  17 \\
    \bottomrule
  \end{tabular}
  \caption{Numbers of remaining prototypes without hurting the  accuracy.}\label{tab:tipping}
\end{minipage}
\end{table}
We perform prototype pruning with varying pruning thresholds to obtain various sizes of remaining prototypes, to analyze how much the pruning impacts the predictive performance.  The idea is, after we obtain a set of prototypes, we compute the frequencies of each prototype being mapped to and then remove prototypes with frequencies lower than a threshold we choose. Then we train a new model with the remaining prototype. We use $\hat{K}$ to represent the number of remaining prototypes.
For demonstration, we choose the Amazon dataset, and we set the pruning threshold to \{0.2\%, 0.5\%, 1\%, 2\%, 5\%, 10\%\}, which returns various models with much fewer prototypes. Their predictive performance and the number of remaining prototypes are reported in Figure \ref{fig:amazon_pruning}. Note that, with only a 0.5\% threshold, 186 prototypes, which is 93\% of the total, were pruned. Meanwhile, removing these prototypes did not hurt the predictive performance at all:  the accuracy after pruning is still 0.918, the same as when the model has 200 prototypes. This implies that a large number of prototypes are redundant and can be safely removed without hurting the model's performance. This brings considerable benefits to interpretability since, after pruning, the model is only left with 14 prototypes with exactly the same accuracy as $K = 200$. This means that in practical uses, a human user, either model designer or end user, can easily check all prototypes to determine whether they make sense, like an example we provide in Section \ref{sec:trajectory}. 

To test whether the observation applies to different initial $K$, we also set $K = 100$ and $K = 400$ and repeat the experiment. The curves are similar to $K = 200$: the accuracy does not change even when a large number of prototypes are pruned. Only when reaching a certain ``tipping point'', the performance starts to drop, then pruning hurts the performance. The results imply that the number of necessary prototypes for a given dataset is more or less the same, even provided with a different number of initial prototypes $K$. The findings above offer meaningful implications for model tuning, that one does not need to tune $K$ heavily: as long as we supply the initial model with a large number of prototypes and let the model achieve the best predictive performance it can obtain, then we can prune the prototypes afterward for better interpretability.%We report the tipping point for different datasets in Table \ref{tab:tipping}. While the tipping point varies for different values of $K$, they all appear around 20 prototypes. 

% \begin{figure}[ht]
%     \centering
%     \includegraphics[width=0.55\linewidth]{figures/pruning.png}
%     \captionof{figure}{The effect of prototype pruning}
%     \label{fig:pruning}
%   \end{figure}

 Therefore, we conduct prototype pruning for all fine-tuned ProtoryNet models from Table \ref{table:acc1} with $K = 200$ and report the minimum $\hat{K}$ that achieves the same accuracy as $K = 200$. $\hat{K}$ for all models are reported in Table \ref{tab:tipping}. Results show that all datasets only require around 20 prototypes, which indicates a significant improvement in interpretability compared to other prototype-based DNNs, given the complexity of the dataset and task. Now model designers or users only need to examine the list of prototypes that could easily fit into a piece of paper, like Table \ref{tab:yelp_prototypes}, to understand or contest the model. 
%     \begin{varwidth}[b]{0.5\linewidth}
%     \centering
% \begin{tabular}{ll}
% \toprule
% \multicolumn{1}{c}{$\theta$} & \multicolumn{1}{c}{$\hat{K}$} \\  \midrule
% 10\%                         & 2                                            \\
% 5\%                          & 5                                            \\
% 2\%                          & 6                                            \\
% 1\%                          & 10                                           \\
% 0.5\%                        & 18                                           \\
% 0.2\%                        & 43       \\ \bottomrule                                   
% \end{tabular}
%     \caption{The number of remaining prototypes at different pruning threshold}
%     \label{tab:Khat}
%   \end{varwidth}%
% \end{table}

% \begin{figure}[h]
% \centering
%  \includegraphics[width=0.45\textwidth]{figures/prototype_pruning.png}
%   \caption{The effect of prototype pruning}\label{fig:k_and_acc}
% \end{figure}

\subsection{Prototypes and Prototype Trajectories}
\label{sec:trajectory}
In this section, we present an example ProtoryNet trained on the Yelp dataset with prototype pruning. As shown in Table \ref{tab:tipping}, this model ends up with only 15 prototypes while maintaining the same accuracy as $K = 200$ (0.962 accuracy as reported in Table \ref{tbl:accuracy}). Table \ref{tab:yelp_prototypes} displays the prototypes along with their corresponding sentiment scores.  The prototypes are ranked in descending order based on their sentiment scores. These prototypes encompass a wide range of sentiments, ranging from the most positive prototype, ``I love this place" with a sentiment score of 0.972, to the least positive prototype, ``I won't be going back" with a sentiment score of 0.011.   It is noteworthy that his highly accurate model only needs prototypes that can fit into half of a page, enabling easy and quick comprehension of all the prototypes.

\begin{table*}[t]
\caption{Prototype information for Yelp review ($\hat{K}$ = 15 after pruning).}
\label{tab:yelp_prototypes}
\centering
\small
\begin{tabular}{l  p{12cm}   c }
    \toprule 
   \textbf{ID} & \textbf{Prototypes}  &  \textbf{Sentiment}   \\
    \midrule
 1 & I love this place & 0.972   \\\hline
 2 &  The biggest breakfast in Pittsburgh, as far as I can tell - and delicious and cheap too  & 0.972  \\\hline
3 & Went here for lunch yesterday with a friend and it was so yummy & 0.968  \\ \hline
 4 & The steak was cooked perfectly and the crabcake was a good size & 0.962  \\ \hline
 5 & Papa J's is by far my favorite restaurant in Pittsburgh, my hometown & 0.949  \\ \hline
 6 & My aunt insisted that we have lunch at Uno's Pizzeria \& Grill as the food was delicious & 0.603  \\ \hline
 7 &From the minute we were seated, we were greeted by a server that was clearly inexperienced and didn't know the menu & 0.171 \\ \hline
 8 & We ended up spending a fortune on beer and mediocre appetizers & 0.074   \\ \hline
 9 &If I want to spend that kind of money, I'll go somewhere that I can get good service & 0.028   \\ \hline
 10 & They finally brought my food out and left it without asking for me to pay  & 0.027  \\ \hline
 11 & The burgers were over cooked and the fries were soggie and the milkshake was runny at best & 0.019  \\ \hline
 12 & The waitress told me that the kitchen hadn't even started on my order yet, so I told her to cancel it and walked out & 0.017 \\\hline
 13 & It took forever to order and then forever and the place was empty & 0.013   \\ \hline
 14 & I won't be going back & 0.011  \\ \hline
         15 & Food was terrible & 0.011  \\ 
    % 16 & I am sorry to say that this was by far the worst experience I have had dealing with anyone & 0.175 \\\hline
    % 17 & I think I won't come back here for a while, especially since the aftermath of the meal was horrible & 0.175 \\ \hline
    % 18 & Service was the worst I've experienced in a long time & 0.175  \\ \hline
    % 19 & I ordered a chicken dish that was the worst tasting meal of my life & 0.175  \\ \hline
    % 20 & The food is mediocre, the staff didn't seem to care, and honestly, we could have gone around the corner to Riva's and gotten better food at their drive-through & 0.175  \\ 
    \bottomrule
  \end{tabular}
\end{table*}

Then, each input text can be represented by a sequence of prototypes selected from Table \ref{tab:yelp_prototypes}. One can regard the sequence of prototypes as a human-understandable representation of the input text. Unlike other sequence encoders based on embedding techniques where the features are not sensible to humans, here we can consider this prototype trajectory as ``prototype encoding'' and the features, i.e., prototypes, are easily understandable.

% \usepackage{multirow}
\renewcommand{\arraystretch}{1.1}
\begin{table}[]
\caption{Two Positive Examples}\label{tab:prototype-trajectory}\label{tab:pos}
\vspace{2mm}
\begin{tabular}{c|p{20em}cl}
\toprule
&\multicolumn{1}{c}{\textbf{Input Text}}                         & \multicolumn{1}{c}{\textbf{Prototype}}                       & \multicolumn{1}{c}{\textbf{Trajectory}} \\ \hline
 \parbox[t]{2mm}{\multirow{7}{*}{\rotatebox[origin=c]{90}{Example 1}}}&\ding{172}This was our first visit to Paradise Bakery and all I can say is YUM                                                    & 3                                                        & \multirow{7}{*}{    \begin{minipage}{.27\textwidth}
      \includegraphics[width=\linewidth]{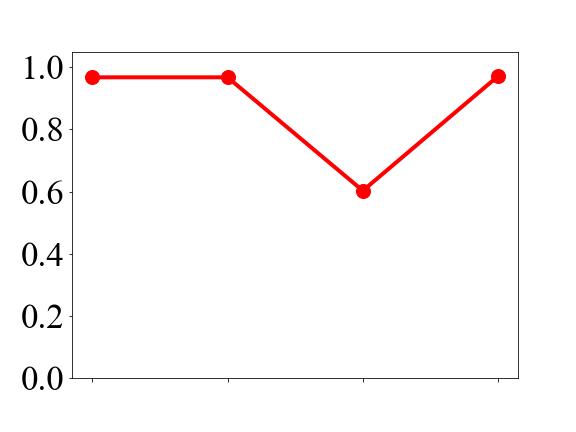}
    \end{minipage}}                       \\ \cline{2-3}
&\ding{173} It was so good, I went back for lunch the next day & 3                               &                                         \\ \cline{2-3}
&\ding{174} The dining room is very pleasant and clean, the service is great and the sandwiches are super yummy                                              & 6                   \\ \cline{2-3}
&\ding{175} I love having this fairly close to our house                                              & 1                                   &                                         \\ \bottomrule
 \parbox[t]{2mm}{\multirow{6}{*}{\rotatebox[origin=c]{90}{Example 2}}}&\ding{172}This place is fantastic   & 1                                                          & \multirow{3}{*}{    \begin{minipage}{.27\textwidth}
      \includegraphics[width=\linewidth]{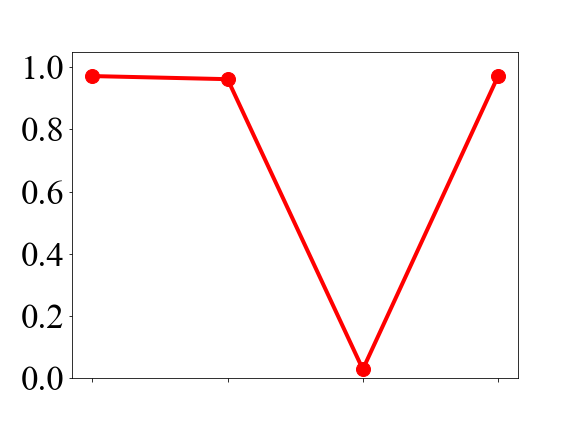}
    \end{minipage}}                       \\ \cline{2-3}
&\ding{173} Impeccable service, great atmosphere and outstanding food & 4                               &                                         \\ \cline{2-3}
&\ding{174}Yes, it's pricey but well worth it.                                               & 9                       \\ \cline{2-3}
&\ding{175}I've been here a couple of times and it never disappoints                                               & 1                                               &                                         \\ \bottomrule
\end{tabular}
\end{table}

 We show two positive examples in Table \ref{tab:pos} and two negative examples in Table \ref{tab:neg}. 
 Each sentence in a text instance is mapped to one of the prototypes from Table \ref{tab:yelp_prototypes} as well as the corresponding sentiment score, generating a trajectory of prototypes and sentiments.  For example, the first sentence in Example 1, ``This is our first visit to Paradise Bakery and all I can say is YUM!'' is mapped to prototype 3, ``Went here for lunch yesterday with a friend and it was so yummy.'' The corresponding sentiment is 0.968, as shown in the sentiment trajectory in the figure. Note that different sentences in the text input can be mapped to the same prototype, such as the second sentence in Example 1: ``It was so good, I went back for lunch the next day'', which is also mapped to prototype 3 in Table \ref{tab:yelp_prototypes}. 

 \renewcommand{\arraystretch}{1.1}
\begin{table}[]
\caption{Two Negative Examples}\label{tab:neg}
\vspace{2mm}
\begin{tabular}{c|p{20em}cl}
\toprule
&\multicolumn{1}{c}{\textbf{Input Text}}                         & \multicolumn{1}{c}{\textbf{Prototype}}                       & \multicolumn{1}{c}{\textbf{Trajectory}} \\ \hline
 \parbox[t]{2mm}{\multirow{5}{*}{\rotatebox[origin=c]{90}{Example 3}}}&\ding{172}I used to LOVE this place  & 1                                                      & \multirow{5}{*}{    \begin{minipage}{.27\textwidth}
      \includegraphics[width=\linewidth]{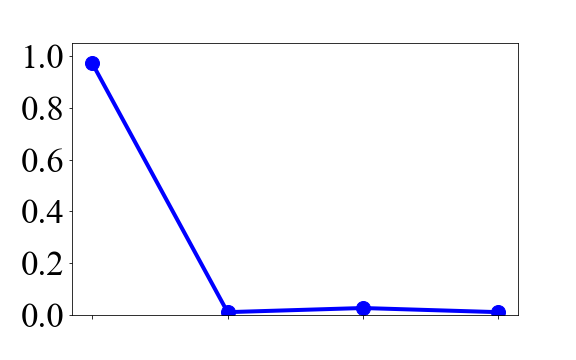}
    \end{minipage}}                       \\ \cline{2-3}
&\ding{173} But the service was TERRIBLE & 15 &                                         \\ \cline{2-3}
&\ding{174}The woman was so slow and put her FINGER in my food                                              & 10                                                      &                                         \\ \cline{2-3}
&\ding{175}I won't be coming back & 14     &                                                                          \\ 
\bottomrule
 \parbox[t]{2mm}{\multirow{5}{*}{\rotatebox[origin=c]{90}{Example 4}}}&
\ding{172}Not good at all                                                         & 15                                                          & \multirow{5}{*}{    \begin{minipage}{.27\textwidth}
      \includegraphics[width=\linewidth]{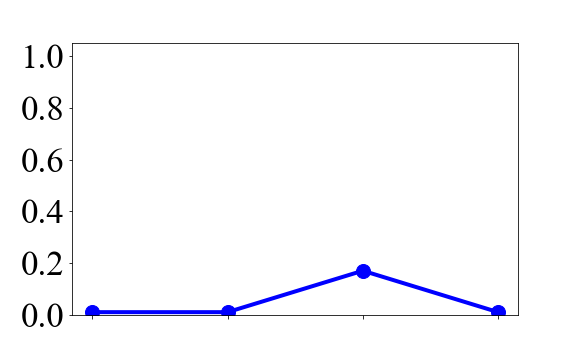}
    \end{minipage}}                       \\ \cline{2-3}
&\ding{173} Average at best & 15                               &                                         \\ \cline{2-3}
&\ding{174} Table we were sat at was sticky and needed wiping down, had to ask the server twice         & 7 & 
\\ \cline{2-3}
&\ding{175} Food was ok but not good        & 15 & \\\bottomrule
\end{tabular}
\end{table}

 We observe that the trajectories can be very different even for the same sentiment class.  Example 1 stays positive for the entire review, while Example 2 starts and ends with positive sentiments but mentions a negative aspect in the middle, that it's pricey. Similarly, the two negative examples also yield different trajectories of sentiments. Example 3 starts with a positive sentiment since the customer ``used to LOVE this place'', which is mapped to prototype ``I love this place'' with a sentiment score of 0.972. Then the customer changes his tone and talks about negative aspects of the restaurant, i.e., bad service and 
unsanitary behavior of the waitress, and ends with a negative sentiment. On the other hand, Example 4 maintains a negative tone for the entire review, which is also reflected by the trajectory of sentiments. As such, the interpretation of ProtoryNet can be more fine-grained, generating deeper insights to users.

 Users can identify a more subtle sentiment development or change of tones in the text that document-level prototypes cannot achieve. From this, users can extract useful information. For example, by identifying the change of tones in positive reviews, the model indirectly teaches restaurants which aspects they should pay attention to and probably improve in the future. For instance, the sentiment trajectory for Example 3 points out that the price might be a little high, and it is something the restaurant needs to take a look at if they would like to increase customer satisfaction. Such information can potentially be more valuable to restaurants than simply predicting whether a review is positive or negative.

\begin{table}[t]
    \centering
    \small
 \caption{Accuracy of ProtoryNet When Substituting LSTM with Other Interpretable Models}\label{tab:substitute}
  \begin{tabular}{l| c ccc}
    \toprule 
   \textbf{ Data set}  & Average &Logistic Regression & Decision Tree & ProtoryNet \\
    \midrule
    IMDB  & 0.602  & 0.859 & 0.836 & 0.914 \\
    Amazon Reviews & 0.559 & 0.808 & 0.878 &0.918\\
    Yelp Reviews & 0.802  & 0.825 & 0.923 &0.962\\
    Rotten Tomatoes & 0.501 & 0.671 & 0.796 &0.881\\
    Hotel Reviews & 0.896 & 0.905 & 0.918 &0.961\\
    Steam Reviews  & 0.771  & 0.815 & 0.790 &0.924\\
    \bottomrule
  \end{tabular}

    \label{tab:other}
\end{table}

\paragraph{ Substituting LSTM with Interpretable Models} The previous analysis demonstrates the diversity in the sentiment trajectory, that even if the predicted sentiments for the whole review are positive (or negative), the trajectories of sentiments could differ greatly from each other. Thus, the trajectory reflects the complexity as well as heterogeneity in the sentiment development along with the text reviews. In the ProtoryNet model, the sentiments are not directly used but implicitly represented by the prototypes. When generating a prediction, a sequence of similarities to the active prototypes is fed to an LSTM model, which is processed by an LSTM model. The LSTM is used to learn the temporal pattern from the sequence to produce the final output. Note that the LSTM is an essential component since other \emph{interpretable} models cannot remember as LSTM does. To demonstrate the value of LSTM, we conduct a set of experiments where we use an interpretable model to replace LSTM in the final step. The features are sentiments of the active prototypes for each sentence in a review. Since the interpretable models work with panel data, we truncate all reviews to 10 sentences and pad those with fewer sentences with the average sentiments from existing sentences. We experimented with two types of interpretable models,  Logistic Regression and Decision Tree. In addition, we calculate the average sentiment of sentences in each review (without padding) and compare it with a threshold to obtain a prediction. Results are shown in Table \ref{tab:substitute} in comparison with ProtoryNet which uses an LSTM to process the sentiment change.

Table \ref{tab:substitute} shows that the original ProtoryNet using LSTM achieves much better performance than the interpretable baselines. The results prove the necessity of using an LSTM that processes the text as a sequence instead of treating it as a collection of sentiments. This means, not only do the sentiment scores matter, but where they appear in the text also matters.

\textcolor{black}{
\vspace{-5mm}
\paragraph{Explaining the Prototype Trajectory Patterns} 
Since LSTM is necessary and cannot be replaced by a simpler interpretable model, we aim to explain the LSTM component. Specifically, we would like to understand what trajectory patterns the LSTM can capture. Explaining RNN/LSTM is always a challenge due to its temporal interactions and non-linear transformations. 
We use the method from the recent work of \citeauthor{hong2022adaax} (\citeyear{hong2022adaax}) as a post-hoc explainer to the LSTM.
This method will generate a  deterministic finite automaton (DFA) that summarizes the patterns an input document needs to match in order to be predicted positively. Here a pattern is a sequence of prototypes. 
For example, if ``7 $\rightarrow$ 4'' is a pattern identified in the DFA, it means a document is positive as long as the first sentence is mapped to prototype 7, the second sentence is mapped to prototype 4, and the following sentences can be mapped to any prototypes since the pattern doesn't specify.  A DFA aggregates the main patterns that exist in the LSTM being explained into a graphical representation. 
See an example trained on Yelp data in Figure \ref{fig:dfa_prototypes_pos} in the appendix. 
This DFA describes the patterns the model used to predict positive sentiment scores for Yelp reviews.
It achieves an explanation fidelity of 91.4\%, meaning that the patterns are correct on 91.4\% of the instances.
Two positive examples from Table \ref{tab:prototype-trajectory} are predicted positive because they match this DFA's patterns (starts with a sentence mapped to  prototype 3 or prototype 1).
Similarly, Figure  \ref{fig:dfa_prototypes_neg} shows an example DFA for predicting negative sentiment; and two negative examples from Table \ref{tab:prototype-trajectory} match this DFA.}

\textcolor{black}{
It is interesting to notice that generally, the model will classify a
review as positive if it stumbles upon a prototype with high sentiment score (prototypes
with low ID in Table \ref{tab:yelp_prototypes}).
Equivalently, the model will classify a review as negative if its first sentence is mapped to a low-score prototype (at the bottom in Table \ref{tab:yelp_prototypes}). 
In addition, using the DFA as an explainer, we can diagnose when LSTM makes mistakes.
For example, if we use 15 prototypes from Table \ref{tab:yelp_prototypes}, the review ``I had the shrimp boil but, it was very under-seasoned. 
The service and atmosphere was great in general. '' will match with  prototypes ``The burgers were over cooked and the fries were soggie and the milkshake was
runny at best.'' (prototype ID 11) and ``I love this place'' (prototype ID 1).
Based on the DFA, this review is predicted as positive by the model since the pattern ``11 $\rightarrow$1'' starts negative and changes the tone to positive. But its true label of the review is actually negative.} 
\subsection{\textcolor{black}{Extension to Multi-class classification }}
\textcolor{black}{ProtoryNet can also be used for multi-label classification.
We only need to modify the fully connected layer (part (d) in Figure \ref{fig:architecture}) to have the number of labels we want.
For illustration, we run experiments using 2 datasets: DBPedia and Consumer complaints.
For each dataset, we extract 4 labels for the multiclass classification tasks and set the number of prototypes to 20.
The dataset is described with further details in the Appendix.}

\textcolor{black}{The prototype label is generated the same way as binary classification: we feed the prototype to the trained model and collect the output label vector, whose dimension is the same as the number of labels.
For example, with 4 labels ``Person'', ``Animal'', ``Building'', ``NaturalPlace'' in dataset DBPedia, prototype ``Eremias acutorostris is a species of Lizard found in east Iran and south Afghanistan'' is represented by a vector (0.019, 0.986, 0.017, 0.014),
where 0.019 indicates how much this prototype belongs to the class ``Person'', etc. 
Because 0.986 is the largest so this prototype is mostly likely to represent ``Animal''.}

\textcolor{black}{
Results show that ProtoryNet still outperforms ProSeNet, reducing the gap compared to the black-box baseline. }

{\color{black}
\begin{table}[h]
    \centering
    \small
 \caption{Accuracy Comparison on Datasets for Multi-class Classification }
 \vspace{5pt}
 \label{tab:substitute}
  \begin{tabular}{l| c ccc}
    \toprule 
   \textbf{ Data set}  & DistilBERT & ProtoryNet & ProSeNet  \\
    \midrule
    DBPedia  & 0.996  & 0.991 & 0.984  \\
    Consumer complaints & 0.967 & 0.927 & 0.878 \\

    \bottomrule
  \end{tabular}

    \label{tab:other}
\end{table}
}
\subsection{Ablation and Sensitivity Analysis}\label{sec:ablation}

In previous experiments, we used the same set of hyper-parameters, to show that our ProtoryNet is easy to use in practice since it does not need heavy tuning and can still achieve reliable performance.
In this section, we evaluate the effect of different hyper-parameters on the model performance. We also examine how much the sparsity transformation hurt the predictive performance, which is designed for better interpretability. 

\begin{figure}[t]
\centering
\begin{minipage}{.5\textwidth}
  \centering
 \includegraphics[width=0.9\textwidth]{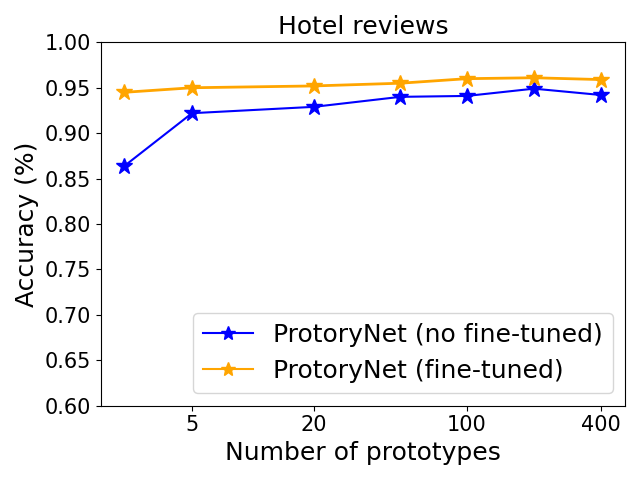}
  \captionof{figure}{Effect of $K$ on accuracy.}\label{fig:k_and_acc}
\end{minipage}%
\begin{minipage}{.5\textwidth}
  \centering
 \includegraphics[trim={1.5cm 0 1.5cm 0},width= 0.75\textwidth]{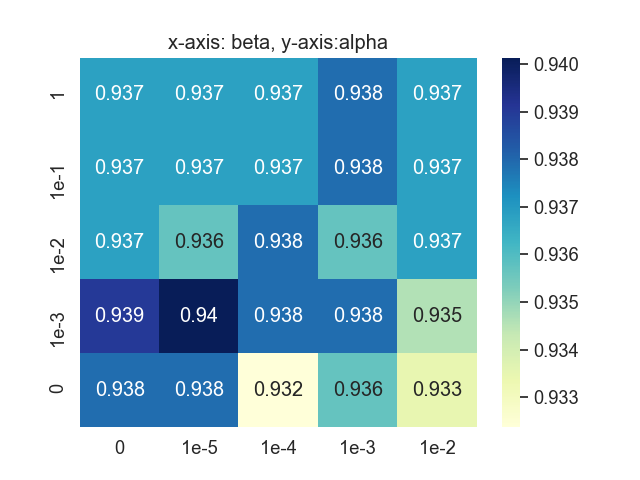}
\captionof{figure}{Sensitivity analysis of $\alpha$ and $\beta$.}\label{fig:sensitivity}
\end{minipage}
\end{figure}

\paragraph{Effect of $K$} We investigated how the initial number of prototypes, $K$, influences the performance of ProtoryNet\footnote{Note that K was selected from [5,20,50,100,200,400] via a validation set when producing Table \ref{table:acc1}.}. In Figure~\ref{fig:k_and_acc}, the performance of ProtoryNet on the Hotel Review data set is plotted for different values of $K$. Other hyperparameters were controlled to be the same.  Curves in Figure \ref{fig:k_and_acc} show that ProtoryNet is not so sensitive to $K$ once $K$ is sufficiently large. This observation can be explained by Table \ref{tab:substitute}, that only a minimal number (\textcolor{black}{about 20}) of effective prototypes are actually needed to ``cover'' the feature space. More prototypes are only redundant for the classification task and can be safely removed.  This finding reinforces the insights for parameter tuning: users just need to set $K$ to a large number and then prune it back. 
% and the performance measures (accuracy) were averaged over 5-fold cross-validation experiments (the whiskers in the figure represent the standard deviation). 
 For the fine-tuned ProtoryNet, the performance is already very well with a small $K$. This is because when fine-tuning is allowed, sentences can be moved toward the prototype they are mapped to, thus it does not need many prototypes to cover the whole space. On the other hand, for non-fine-tuned ProtoryNet, each sentence is represented by a fixed vector in the feature space. If there are very few prototypes, it becomes difficult for some sentences to be mapped to the correct prototype since they are far away from all prototypes.

\paragraph{Effect of Diversity and Prototypicality Terms}
 We performed a sensitivity analysis to understand the effect of the two terms on predictive performance using the Hotel dataset. Since our goal was to study the effect of $\alpha$ and $\beta$, we fixed the $K$ to be 100 and tried different combinations of $\alpha, \beta$, where $\alpha = 0, 1e^{-3},1e^{-2},1e^{-1}, 1,$ and $\beta = 0, 1e^{-5}, 1e^{-4}, 1e^{-3}, 1e^{-2}$. As seen in Figure~\ref{fig:sensitivity}, our experiment revealed that the ProtoryNet achieves consistently high performance with different $\alpha$ and $\beta$. This benefits the training and tuning process because users do not need to invest a tremendous amount in parameter tuning. In this paper, we set $\alpha = 0.1$ and $\beta = 1e^{-4}$ to all experiments. %ProtoryNet was more sensitive to $\beta$, which may reveal the trade-off between the different loss terms.
% The results also illustrate how different loss terms impact the overall performance - when $\alpha = 0$ or $\beta = 0$. %Interestingly, when both $\alpha$ and $\beta$ were set to zero (i.e. only the accuracy term alive in the loss function), the accuracy of ProtoryNet was not the highest, which was against our initial expectation.
In addition, we notice that the best performance was achieved when $\alpha$ and $\beta$ are set to small values instead of 0, suggesting the positive impact of the diversity term and prototypicality term we designed on the predictive performance. A possible explanation would be that having some constraints on the prototypes' diversity ($\mathcal{L}_\text{div}$) and their representativeness ($\mathcal{L}_\text{proto}$) %indeed yield some positive effects on the performance of prediction 
prevents overfitting as these terms ``regulate'' prototypes.  
% Sensitivity analysis for other datasets are included in the supplementary material.

\paragraph{Effect of Sparsity Transformation} Furthermore, we conducted an ablation study on the sparsity transformation. The sparsity transformation from $\tilde{\mathbf{S}}$ to ${\mathbf{S}}$ was used to enhance the interpretability of the model, which forces each sentence to be mapped to one closest prototype, i.e., the \emph{active prototype}. Without the sparsity transformation, each sentence will be mapped to $K$ prototypes, which will involve $T\cdot K$ prototypes in the explanation for the prediction. Despite the big advantage of sparsity transformation in interpretability, we investigate the impact of this sparsity transformation on predictive performance. 
We measured the change in prediction accuracy when the sparsity transformation step had been removed, and the dense similarity matrix $\tilde{\mathbf{S}}$ had been used directly.  Specifically, we compare fine-tuned ProtoryNet's performance with and without the sparsity transformation and show the comparison in Table \ref{tbl:sparse_dense}. %The result revealed that there was only a small drop of accuracy (approx. 1\%) caused by the sparsity transformation, while the benefit of interpretability was huge (see the supplementary material for the detail). 
As reported in Table~\ref{tbl:sparse_dense}, there was only a small drop in accuracy (approx. 1\%) caused by the sparsity transformation.
\vspace{-3mm}
\begin{table}[h]
\small
\caption{Performance comparison between non-sparse $\tilde{\mathbf{S}}$ and sparse $\mathbf{S}$ as the input to the LSTM layer.
The validation accuracy for each case. }
\label{tbl:sparse_dense}
\centering
  \begin{tabular}{l| c c}
    \toprule 
   \textbf{ Data set}  & \textbf{Dense (K active prototypes)} & \textbf{Sparse (1 active prototype)}\\
    \midrule
    IMDB  & 0.920  & 0.914 \\
    Amazon Reviews &  0.921 & 0.918 \\
    Yelp Reviews & 0.956 & 0.954  \\
    Rotten Tomatoes & 0.896 &  0.881 \\
    Hotel Reviews & 0.968 & 0.961 \\
    Steam Reviews & 0.936 & 0.924\\
    \bottomrule
  \end{tabular}
%   \label{table:acc3}
\end{table}
% \textcolor{magenta}

% {things to do}
% \begin{enumerate}
%     \item use cosine (later)
%     \item remove argmin and use a non-sparse similarity vector (now! important!)
%     \item test the sensitivity of $\eta$ (later)
% \end{enumerate}

% \newpage
% \clearpage

\section{Human Evaluation of ProtoryNet}\label{sec:human}
\textcolor{black}{
In this section, we evaluate the interpretability of ProtoryNet via human evaluations. 
To this end, we designed two surveys.
The first survey evaluated whether individual prototypes picked by the models match the human users' expectations and how easily they can be interpreted.
The second survey evaluated  whether users understood the prototype trajectories at the document level.}

\subsection{Survey 1: Prototype Evaluation}
The first survey evaluated the interpretability of prototypes selected by ProtoryNet. 
We collected responses from 111 individuals, among which 42 identified themselves as non-technical users. 
Subjects were recruited through two different channels. Individuals from the authors' home institution holding a master's degree or above having advanced knowledge of RNNs have been recruited as technical users. Non-technical users were recruited from Amazon Mechanical Turk. 
The survey designs are disclosed in Appendix.

% \textcolor{red}{add 2 survey sub-sections}

% In the survey, we aimed to answer the following questions: 1) How intelligible is a prototype-trajectory explanation of a paragraph?; and 2) How does the interpretability vary between technical users and non-technical users?

\begin{figure*}[ht]
\centering
 \includegraphics[width=\textwidth]{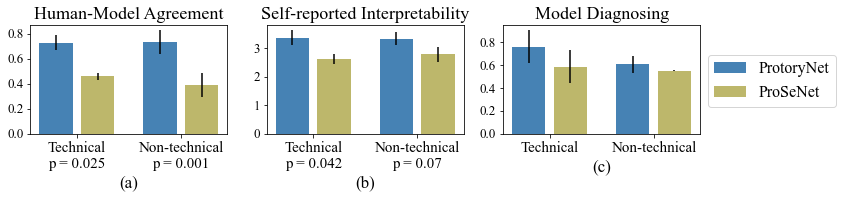}\caption{The means and standard errors (error bars) of the rating of users.  The p-values for the t-test are evaluated for comparing the responses for ProtoryNet and ProSeNet on technical users and non-technical users, respectively.
 }\label{fig:survey}
\end{figure*}

We first evaluated whether the prototypes were indeed representative of the input text to the human users. We asked the users to choose the most appropriate prototype for a given sentence out of four options presented to them, one of which was the actual prototype matched by the model, the other two were randomly selected from the rest of the prototypes, and the other was ``None of the above.'' We created 10 such questions by sampling reviews from the Yelp Review data set, each for ProtoryNet and ProSeNet.
As reported in Figure~\ref{fig:survey}a, ProtoryNet showed a more significant agreement between the model-selected prototype and the prototype that the human users found the most appropriate. For both technical users and non-technical users, ProtoryNet was significantly better than ProSeNet, as was validated by the t-test. The difference between technical users and non-technical users was insignificant, suggesting that non-technical users can comprehend prototypes equally well as technical users.

The survey also included self-report questions to assess how easy it was for them to select a prototype in a score ranging between 1 (very difficult) and 5 (very easy). As reported in Figure \ref{fig:survey}b, subjects found that ProtoryNet was easier to interpret in general,  and the improvement in interpretability was more significant for technical users. %Again, the difference between technical users and non-technical users was insignificant.

Second, we measured how easily the users can learn to interpret the prototype-based explanations from ProtoryNet and ProSeNet. For this, each subject was randomly assigned to either ProtoryNet or ProSeNet and trained on how the model that they are assigned to makes predictions. Then, their proficiency was measured by showing them three examples on which the model had made an incorrect prediction and asking them to diagnose the problem by pointing out an inappropriately matched prototype. The problematic prototype (\textit{i.e.}, the ``correct answer'' for the survey question) was determined via a discussion among the authors, which later turned out to be aligned with the consensus in the survey responses as well. As in Figure \ref{fig:survey}c, both subject groups were more accurate at diagnosing ProtoryNet in general. 
We notice that while technical users find ProtoryNet easier to diagnose, such a difference was not significant for non-technical users. In fact, there was no significant difference between technical users and non-technical users when they use ProSeNet since it was almost equally difficult for these two groups of users.

The reason that ProtoryNet is generally easier to understand than ProSeNet is that the prototypes are defined at the sentence level so it is easier for users to compare and relate a sentence to a prototype. For ProSeNet, however, prototypes are defined at the document level. When a document is long, it may discuss aspects and contain mixed sentiments, thus harder for users to find similar prototypes. 

% \textcolor{black}{We hypothesize that because ProtoryNet's prototypes are at sentence-level, making it shorter and easier to interpret than ProseNet's prototypes, which are at document-level.}

% \textcolor{red}{discussion: why is ProtoryNet is more interpretable}

\subsection{Survey 2: Prototype Trajectory Evaluation}

The second survey evaluated the understandability of explanations provided by a ProtoryNet at the document level, i.e., the trajectory of prototypes. We asked participants to choose the correct prototype trajectory for a given review out of four different prototype trajectories we created from the models' prototypes. An example of the questions is shown in Figure \ref{fig:survey_trajectory} in the Appendix.
If the users can choose the correct trajectory, it means they not only understand the prototypes but also the trajectory of prototypes.

In addition, we also investigated whether different numbers of prototypes $K$ will affect human interpretation. With a small number of prototypes, the distances between a sentence and the closest prototype increase, so their similarities become less apparent to users.  On the other hand, the prototypes tend to be more distinctive, and it is easier for users to select from fewer options. Therefore, it is interesting to investigate the impact of $K$ on choosing the correct trajectory. %On the other hand, with a higher number of prototypes, the distances between prototypes in the latent space decrease, and the differences between prototypical sentences become too subtle. Thus, it is harder to select the correct prototype for a given sentence from 100 prototypes than from only 15 prototypes.
To this end, we trained two ProtoryNet models on the Yelp dataset, one with 15 prototypes (reported in Table \ref{tab:yelp_prototypes}) and the other with 100 prototypes. We then asked users to choose the trajectory for the 4 examples in Table \ref{tab:prototype-trajectory} and \ref{tab:neg} and create four questions for each model. A user is then randomly assigned to see one set of questions for either $K = 15$ or $K = 100$. 

\begin{figure*}[ht]
\centering
 \includegraphics[width=0.5\textwidth]{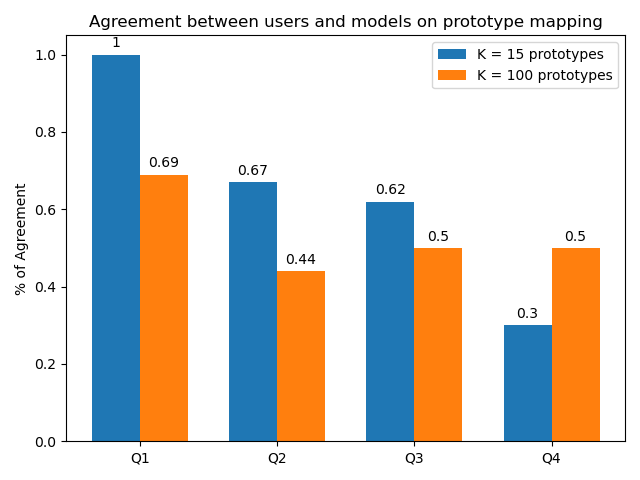}\caption{Agreement between users and the models. 
 Here we have 2 models, one is trained with $K = 15$ prototypes and another is trained with $K = 100$ prototypes.
 The y-axis shows the percentage of correct answers (the users picked the same mapping as the model). 
 The x-axis indicates the survey question.}\label{fig:user_agreement}
\end{figure*}

After filtering out users who had incomplete answers or spent too little time (less than 60 seconds on the four questions), we kept responses from 37 users and reported the accuracy on each of the four questions in Figure \ref{fig:user_agreement}, for the two ProtoryNet models, respectively. The average accuracy across users and questions is around 60\%, which is lower than users' accuracy in selecting prototypes in the previous survey. This is because choosing the correct trajectory includes understanding multiple prototypes and their corresponding sentiments and, thus, is more difficult for users than only selecting individual prototypes. % when the users only needed to map sentences to prototypes, since mapping an entire trajectory of prototypes increases the difficulty. 

% \textcolor{black}{In addition, we measure whether increasing the number of prototypes from $K = 15$ to $K = 100$ make the selection more difficult. 
% To achieve this, we conducted a two-sample t-test from user answers.
% Each sample is constructed by collecting the user's percentage of agreement from each model (one data point is one user's percentage of agreement).
% The results show that the model with $K=15$ has a higher user agreement than the model with $K=100$ (average user agreement 64\% and 53\% respectively), but it is not statistically  significant (p-value = 0.262).
% % We hypothesize that the smaller number of prototypes, the more distinctive the prototypes are and it was easier for humans to select the correct prototype for each sentence.
% }

% \textcolor{black}{p value?}

\section{Discussion and Conclusion}\label{sec:conclusion}
We introduced a novel idea of prototype trajectory in DNNs. Our model, ProtoryNet, maps a text input into a sequence of prototypical sentences, illuminating the underlying dynamics of semantics within the text data. %In addition, sentence-level prototype trajectory allowed fine-grained analysis of paragraphs.  
Therefore, Users can identify a more subtle sentiment development or change of tones in the text that document-level prototypes cannot achieve.
ProtoryNet achieved a predictive performance higher than the state-of-the-art interpretable baselines and reduced the performance gap compared to black-box DNNs.  Moreover, the human evaluation result suggested that ProtoryNet provided more intuitive prototypes than the baseline and that the novice users were able to interpret ProtoryNet equally well as the expert users. The prototype pruning we design has proved to be quite effective on all datasets we experimented with and the resulting models only need around 20 prototypes in total, which is a significant improvement compared to other baselines. 

Our model has also shown be very easy to use. First, it does not rely on heavy parameter tuning (we used the same set of parameters in all of our experiments), which makes it convenient in practice.
In addition, we experimented with two versions of ProtoryNet, a fine-tuned ProtoryNet to fully utilize the power of a state-of-the-art Transformer encoder, and a non-fine-tuned ProtoryNet, which is much smaller, lighter, and energy-efficient. Results show that even the non-fine-tuned ProtoryNet already beats the interpretable baselines and can potentially be more promising with the increasing need for Green AI.

The benefit of prototype-based reasoning resides in the fact that it hides technical details by encapsulating them with prototypical examples while still being tractable numerically when desired. Hence, novice users can understand how the reasoning was achieved in RNNs so long as they can comprehend the prototypes, lowering the barrier for those numerous non-technical users who may use RNN-based applications in the real world. On the other hand, numerical weights assigned to prototypes alongside their association with the ``nuts and bolts'' of RNNs still allow experts to perform in-depth analyses of how a model has drawn a prediction. One can think of the prototypes as a special type of \emph{feature representation} of the original text input. Compared to other types of latent features produced by complicated transformations through encoding layers, where the features are not sensible, the ``prototype encoding'' in ProtoryNet obtains a human-understandable feature representation. Prototypes make sense to humans while encapsulating all necessary information in them, thus they are able to obtain good predictive performance even using only the LSTM layers to process them.

ProtoryNet can potentially be applied to other sequence data other than text. However, one should be able to define meaningful and consistent sub-sequences, like sentences in a document. This definition is task-specific and may need to conform with application-specific constraints. In addition, one may remove the sentence encoder or replace it with some other feature extractor.
% Our immediate future work would be to apply ProtoryNet to other types of sequence data, such as medical data, longitudinal data, etc. In a more theoretical context,

\paragraph{Limitations and Future Work}
In ProtoryNet,  the mapping to the prototype with similar meanings depends on the quality of embeddings. We used Google Universal Encoder, which performs well in most cases but there exist some cases where two sentences are close to each other but with different meanings. Now with the recent breakthrough of ChatGPT and a series of on-going effort in developing LLM, we believe this problem will be less of an issue eventually. 
In addition, the similarity score between a sentence and a prototype is purely based on the embedding, thus it does not naturally have an explanation. Users may still not understand why one sentence  is similar to another. We believe there's an opportunity for future research to rationalize the similarity score computation, especially if using LLM such as GPT-4.
In addition, for future work, it would be interesting to mathematically formalize some of the well-established requirements to be a prototype in the linguistics literature. For example, Panther and K\"{o}pcke~\citep{panther2008} assert several conditions that a prototype must possess---a prototypical sentence is an affirmative declarative sentence; the subject is in the nominative case; the verb in a prototype is in the active voice and in the indicative mood; to list a few. Albeit non-trivial, the mathematical translation of such conditions should bring more interpretability and, perhaps, better performance of ProtoryNet.

% \textcolor{red}{add regularization to the prototypes}

% \section*{Broader Impact}
% %instructions:  order to provide a balanced perspective, authors are required to include a statement of the potential broader impact of their work, including its ethical aspects and future societal consequences. Authors should take care to discuss both positive and negative outcomes.
% The immediate impact of this paper would be the enhanced interpretability of RNNs via the novel idea of prototype trajectory. We made a step towards bridging the prototype theory in linguistics with the interpretability research in deep learning, which could promote multidisciplinary research efforts. Another positive impact of the present work could be resulted from the fact that it targeted a broader group of users, even the ones with less or no technical backgrounds. Hence, we anticipate that more people could benefit from such an interpretable model and, consequently, that the performance of RNN models could become clearer and more transparent to public.

% On the other hand, the negative impact may include the decreased accuracy at the expense of interpretability. It was clear from our experiments that the state-of-the-art black box model outperformed interpretable models by margin. In addition, users of ProtoryNet should be cautious that a prototype is just the closest example to a given sentence, not necessarily an example that carries the same meaning. Hence, users should not equate prototypes with actual sentences when they analyze the results.

\bibliography{JMLR_camera}
% \newpage
\appendix
\section{Reproducibility}
\subsection{Data Sets Description}\label{sec:datasets}
\paragraph{IMDB Movie Reviews} The IMDB Movie Reviews data set is a standard benchmark data set for binary sentiment classification and is available at \url{https://ai.stanford.edu/~amaas/data/sentiment/}.
The data set is perfectly balanced and comprised of 25,000 movie reviews for training and 25,000 for testings and we followed this original partition of the training and testing set and use 10\% of the training data as validation.

\paragraph{Yelp Reviews} The Yelp Reviews data set was
obtained from \url{http://goo.gl/JyCnZq}.
The data set is comprised of 580,000 Yelp review samples and their corresponding labels. 
The authors of the data set have binarized the sentiment scores by assuming 1 and 2 stars as a negative sentiment and 3 and 4 stars as a positive sentiment.
They also already split the dataset into a training set with 550,000 reviews and a test set with 30,000 reviews.
In this paper, we followed this data partition and partitioned the training set into 90\% training and 10\% validation.

\paragraph{Amazon Product Reviews} 
Similar to the Yelp Reviews dataset, we also obtained Amazon Reviews from \url{http://goo.gl/JyCnZq}.
For this dataset, we took random samples of 30,000 reviews, in which 24,000 reviews are randomly selected as the training set and validation, and the remaining 6,000 reviews are used as the test set.

\paragraph{Rotten Tomatoes}
The Rotten Tomatoes Movie Review data set is a corpus of movie reviews used for sentiment analysis and is available at \url{https://github.com/nicolas-gervais/rotten-tomatoes-dataset,} which contains 480,000 reviews.
We randomly split the dataset into a training set and a test set by a ratio of 80-20. Then 10\% of the training data were used as validation.

\paragraph{Hotel Reviews}
The Hotel Reviews data set is comprised of 20,000 review samples evaluating 1,000 hotels and is available on Kaggle: \url{https://www.kaggle.com/datafiniti/hotel-reviews}.
In this paper, we assumed a positive sentiment for reviews of 4 and 5-star ratings and a negative sentiment for reviews of 1 and 2 stars. 
Reviews with 3 stars were ignored. This assignment yields 17,746 positive reviews and 2,254 negative ones. To balance out the data set, we randomly picked 2,254 positive reviews to make them equal, making the total of 4,508 reviews used in our experiments.

\paragraph{Steam Reviews}
The dataset contains reviews from Steam's best-selling games as of February 2019 and is available on Kaggle \url{https://www.kaggle.com/luthfim/steam-reviews-dataset}.
We preprocessed the data by removing potentially incomplete reviews (with less than 10 characters or 2 sentences) and sampling 65,000 positive reviews and 65,000 negative ones.

\paragraph{DBPedia}
This is a multiclass dataset extracted from information on Wikipedia.
The dataset is always maintained up-to-date on \url{http://wikidata.dbpedia.org/develop/datasets}.
For the experiments in this paper, we only use 4 labels ``Person'',``Animal'',``Building'' and ``NaturalPlace''.
\paragraph{Consumer complaints}
The dataset is available on \url{https://www.kaggle.com/datasets/dushyantv/consumer_complaints}.
This is also a multiclass dataset.
For the experiments, we only use 4 classes ``Checking or savings account'', ``Credit card or prepaid card'',``Debt collection'',``Mortgage''.
% \textbf{AG News:} The AG News dataset \footnote {} consists of news articles from the AG’s corpus of news articles on the web pertaining to the 4 largest classes with 30k articles for each class. 

% \textbf{20 Newsgroups: }The 20 Newsgroups data set is a benchmark dataset for many multi-label text classification tasks. It is a collection of approximately 20,000 newsgroup documents, partitioned (nearly) evenly across 20 different newsgroups. 

For pre-processing, the period (`.'), the question mark (`?'), and the exclamation mark (`!') were used as delimiters to define the boundary between sentences. All words were then converted to the lowercase and punctuations were removed using the definition in \texttt{string.punctuation} constant in Python 3.5.
In all experiments, we used pre-trained BERT-based language model with mean-tokens pooling \cite{reimers2019sentence} to convert the raw sentence data to sentence embeddings.
\subsection{Models}\label{sec:models}

\paragraph{Vanilla LSTM} We used 300-dimensional GloVe word embeddings \cite{pennington2014glove} to encode words in sentences. An LSTM model with 2 hidden layers of size 128 each was used. The final prediction was made by a fully connected layer of size 256. A dropout layer of the rate 0.5 was used immediately before the fully connected layer.
The implementation is done in Tensorflow 1.15.

\paragraph{DistilBERT} DistilBERT \cite{sanh2019distilbert} is considered as a lightweight version of the state-of-the-art BERT model with smaller, faster, and less expensive deployment time and resources. In our experiments, a pre-trained DistilBERT model was transferred and fine-tuned to each target data set.
We used an implementation that was available in the Hugging Face Transformers Library (\url{https://github.com/huggingface/transformers}), which was implemented in PyTorch and TensorFlow 2.0.

\paragraph{ProSeNet} ProSeNet \cite{ming2019interpretable} is a state-of-the-art prototype-based interpretable RNN. 
For the implementation of ProSeNet, we used an LSTM layer with 2 hidden layers of size 128 and the dropout rate 0.5 for the sequence encoder.
This is the same configuration as ProtoryNet's RNN layer. 
% We fixed $K = 200$ for all experiments for both ProSeNet and ProtoryNet to generate results for Table \ref{table:acc1}. 
% For fair comparison, we used the fixed constant $K = 200$ for both ProtoryNet and ProSeNet.
We tuned K from $[5,20,50,100,200,400]$ using a validation set.

\paragraph{ProtoryNet}
We used TensorFlow v2.3\footnote{https://www.tensorflow.org/}
to implement ProtoryNet (and v1.15 for other benchmark models).
% As mentioned in the main paper, we fixed $\alpha = 0.1$ and $\beta = 1e^{-4}$ to all experiments.
In addition, the LSTM layer in ProtoryNet was implemented to have the same architecture as the baseline methods to eliminate the bias.
Just like ProSeNet, we tuned $K$ from $[5,20,50,100,200,400]$ using a validation set and fixed $\alpha = 0.1$ and $\beta = 1e^{-4}$.

\paragraph{Bag-of-words}
We followed the ``Bag-of-words and its TFIDF'' in Section 3.1 in paper \cite{zhang2015character}. 
While being considered traditional, the method still achieved very good performance in many cases.
We use TFIDF (term-frequency inverse-document-frequency)
as the word counts, and Logistic Regression as the classifier for the purpose of interpretability.
The method is implemented in Python and Scikit-learn libraries with default configuration.

% \section{List of prototypes in Section 4.3}
% Figure \ref{fig:trajectory1} to Figure \ref{fig:trajectory4} show the prototypes and their sentiment scores used in the main text Section 4.3 and the main text Figure 4. 
% Note that several sentences could be mapped to one prototype.
% \begin{figure}[h]
%  \includegraphics[width=\linewidth]{figures/p1.PNG}\caption{Prototype trajectory 1.}\label{fig:trajectory1}
% \end{figure}
% \begin{figure}[h!]
%  \includegraphics[width=\linewidth]{figures/p4.PNG}\caption{Prototype trajectory 2.}\label{fig:trajectory2}
% \end{figure}
% \begin{figure}[h!]
%  \includegraphics[width=\linewidth]{figures/p3.PNG}\caption{Prototype trajectory 3.}\label{fig:trajectory3}
% \end{figure}
% \begin{figure}[h!]
% % \vspace*{0.25cm}
%  \includegraphics[width=\linewidth]{figures/p5.PNG}\caption{Prototype trajectory 4.}\label{fig:trajectory4}
% \end{figure}

\section{\textcolor{black}{Supplementary Figures}}

This section provides some supplementary figures used in the main paper.

  \begin{figure}[h!]
  \begin{center}
  \vspace{-25mm}
 \includegraphics[width=1.07\linewidth]{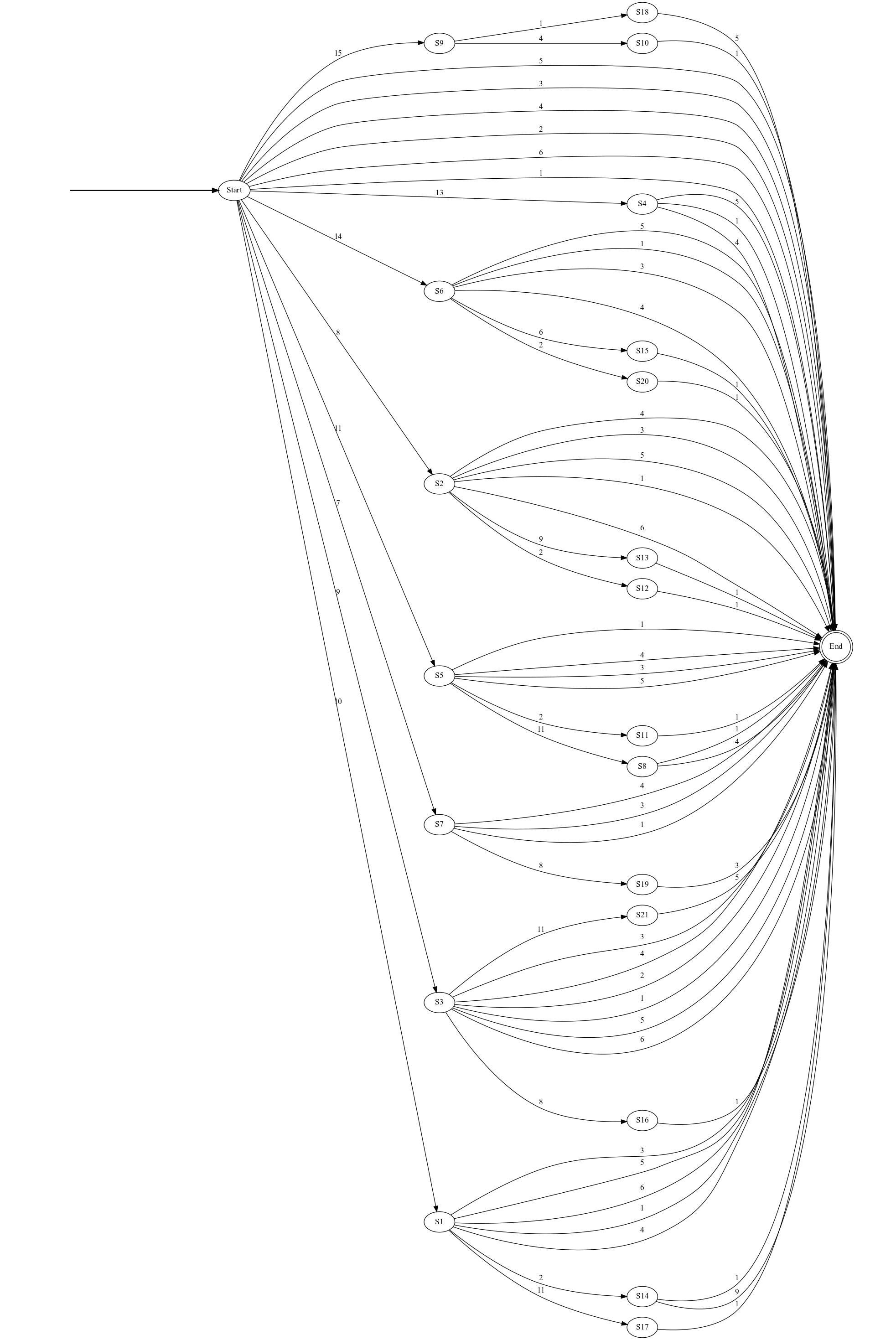}
 \caption{\textcolor{black}{A deterministic finite automaton (DFA) explaining how the LSTM makes decision on predicting if a  review's sentiment is positive.
 The arrow are the prototype IDs from Table \ref{tab:yelp_prototypes} and the ovals are transition states.}
 }
 \label{fig:dfa_prototypes_pos}
 \end{center}

\end{figure}

 \begin{figure}[h!]
  \begin{center}
 \includegraphics[width=0.97\linewidth]{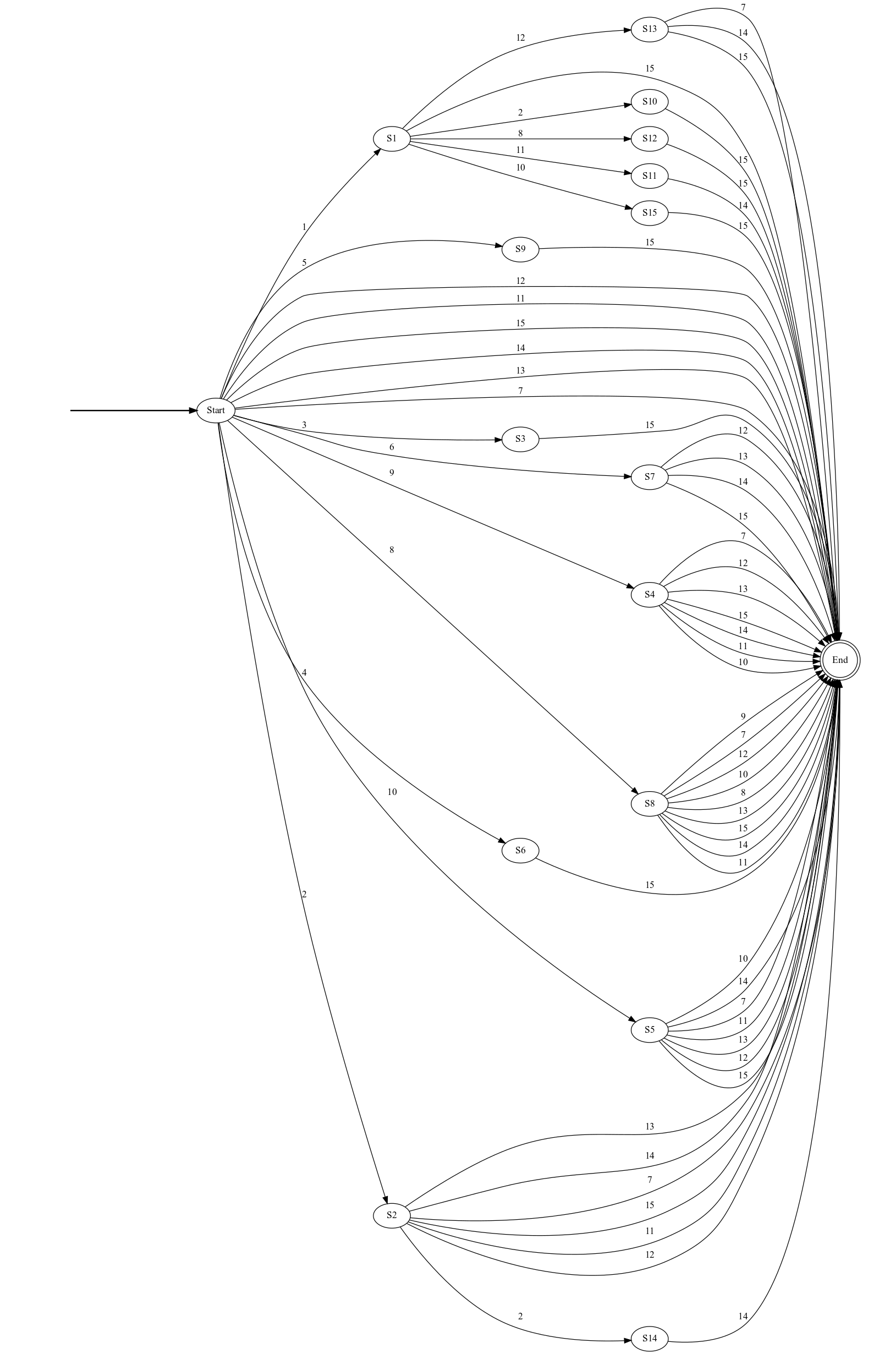}
 \caption{\textcolor{black}{A DFA explaining how the LSTM makes decisions on predicting if a  review's sentiment is negative.
 Similar to Figure \ref{fig:dfa_prototypes_pos}, the arrow are the prototype IDs from Table \ref{tab:yelp_prototypes} and the ovals are transition states.}
 }
 \label{fig:dfa_prototypes_neg}
 \end{center}

\end{figure}  

\section{Survey Questions}
The figures below show a few examples of the survey questions we used for the user evaluation study.

\begin{figure}[h!]
 \includegraphics[width=\linewidth]{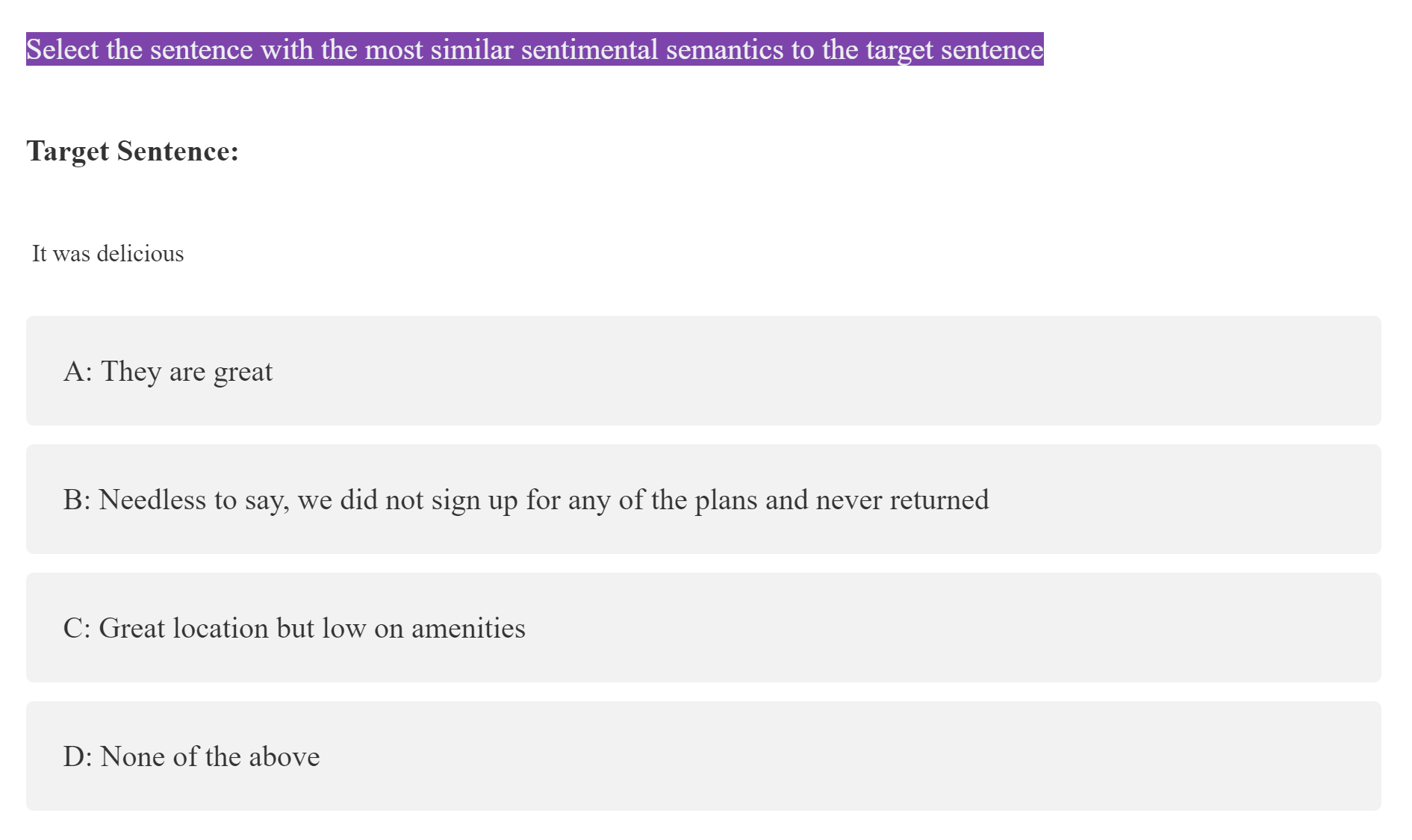}\caption{Prototype selection question.}\label{fig:survey_prototype}
\end{figure}

For the prototype selection, we created 10 questions each, for ProSeNet and ProtoryNet. Here we only show one example in Figure \ref{fig:survey_prototype}.

\begin{figure}[h!]
 \includegraphics[width=\linewidth]{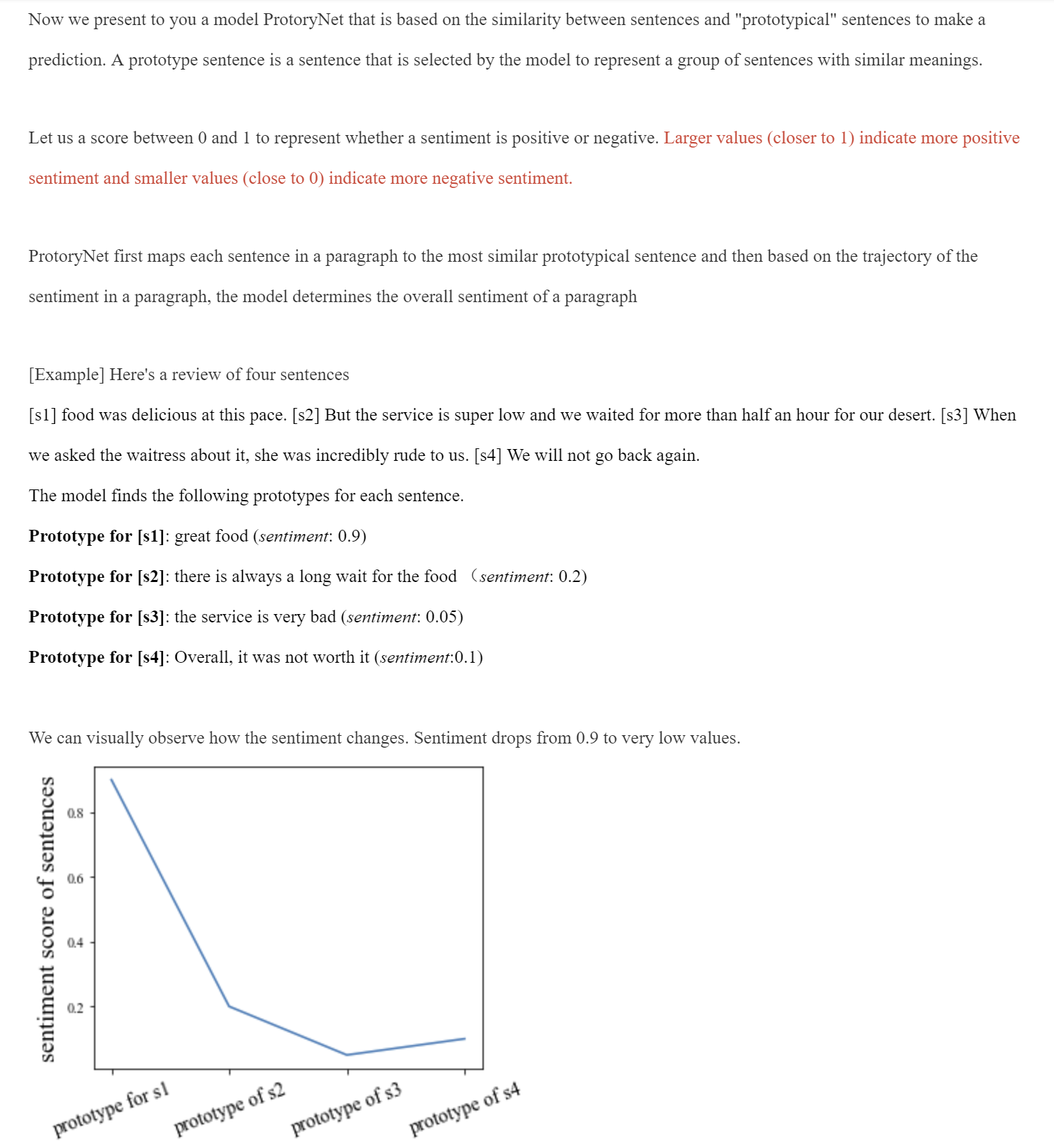}\caption{Education material for ProtoryNet}\label{fig:survey_ProtoryNet_education}
\end{figure}

\begin{figure}[h!]
 \includegraphics[width=\linewidth]{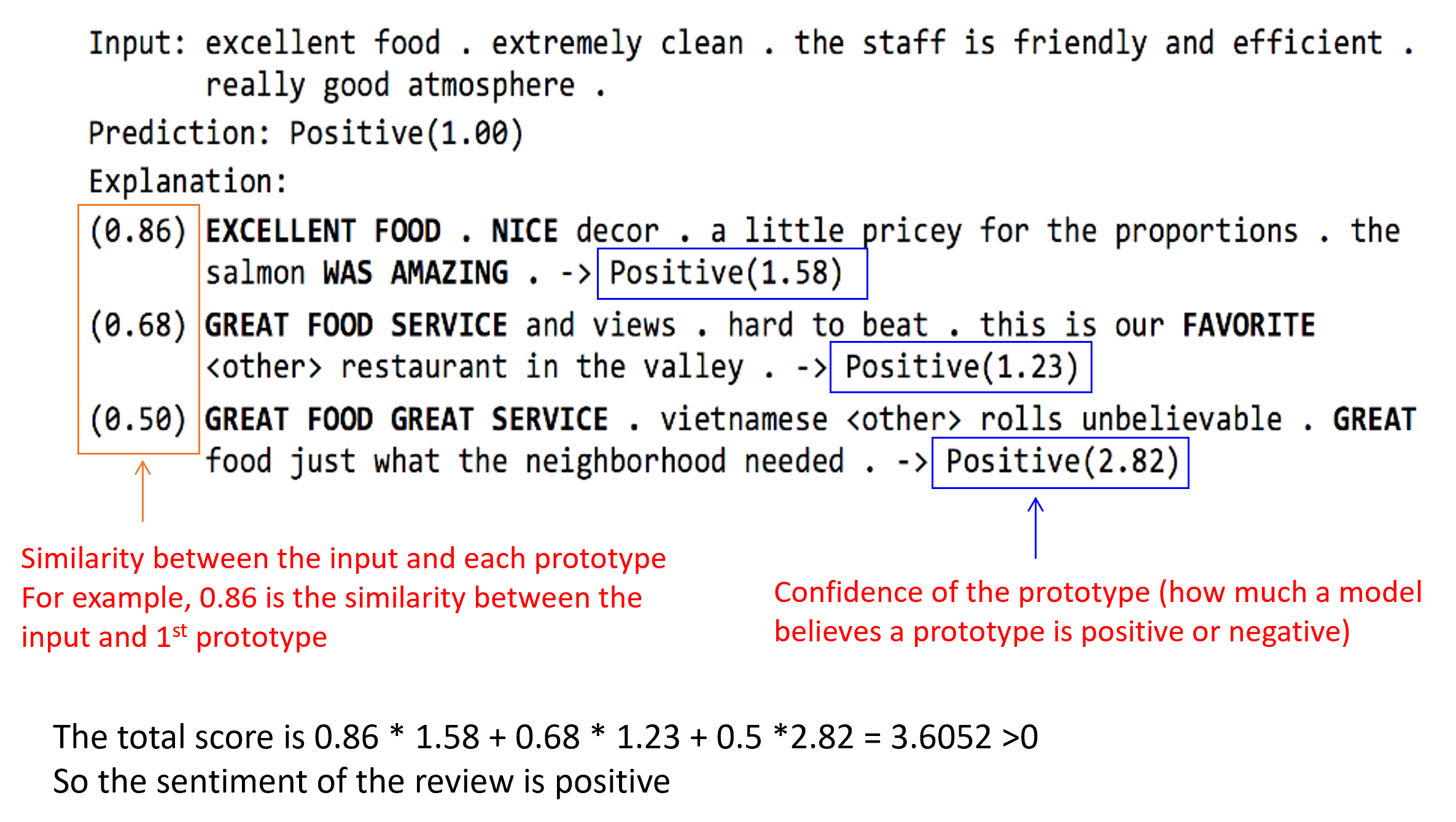}\caption{Education material for ProSeNet.}\label{fig:survey_education}
\end{figure}

Figure \ref{fig:survey_ProtoryNet_education} and Figure \ref{fig:survey_education} show how we educated the subjects about how ProtoryNet or ProSeNet work.

\begin{figure}[h!]
 \includegraphics[width=\linewidth]{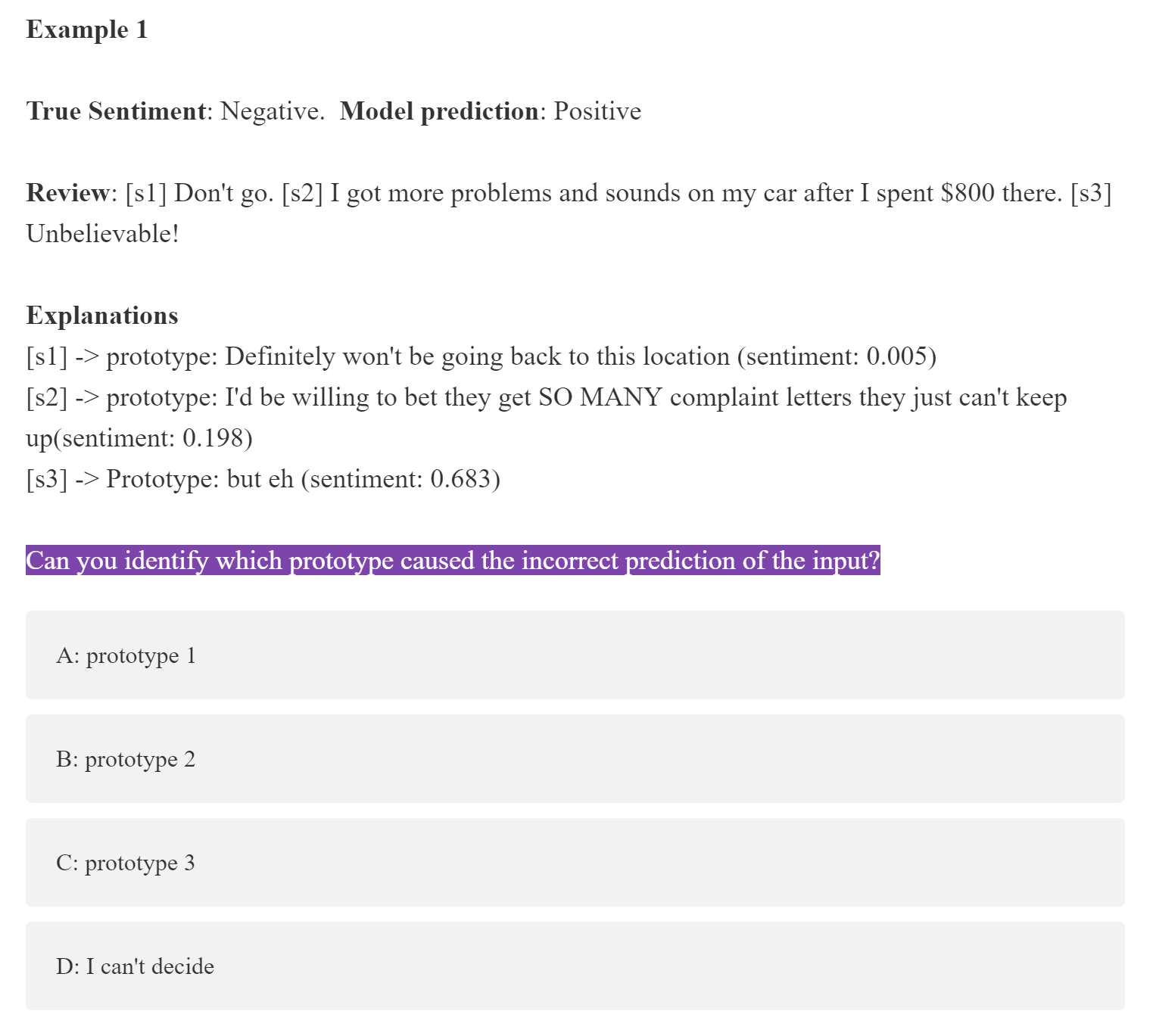}\caption{Diagnosis question for ProtoryNet model.}\label{fig:survey_ProtoryNet_diagnose}
\end{figure}

\begin{figure}[h!]
 \includegraphics[width=\linewidth]{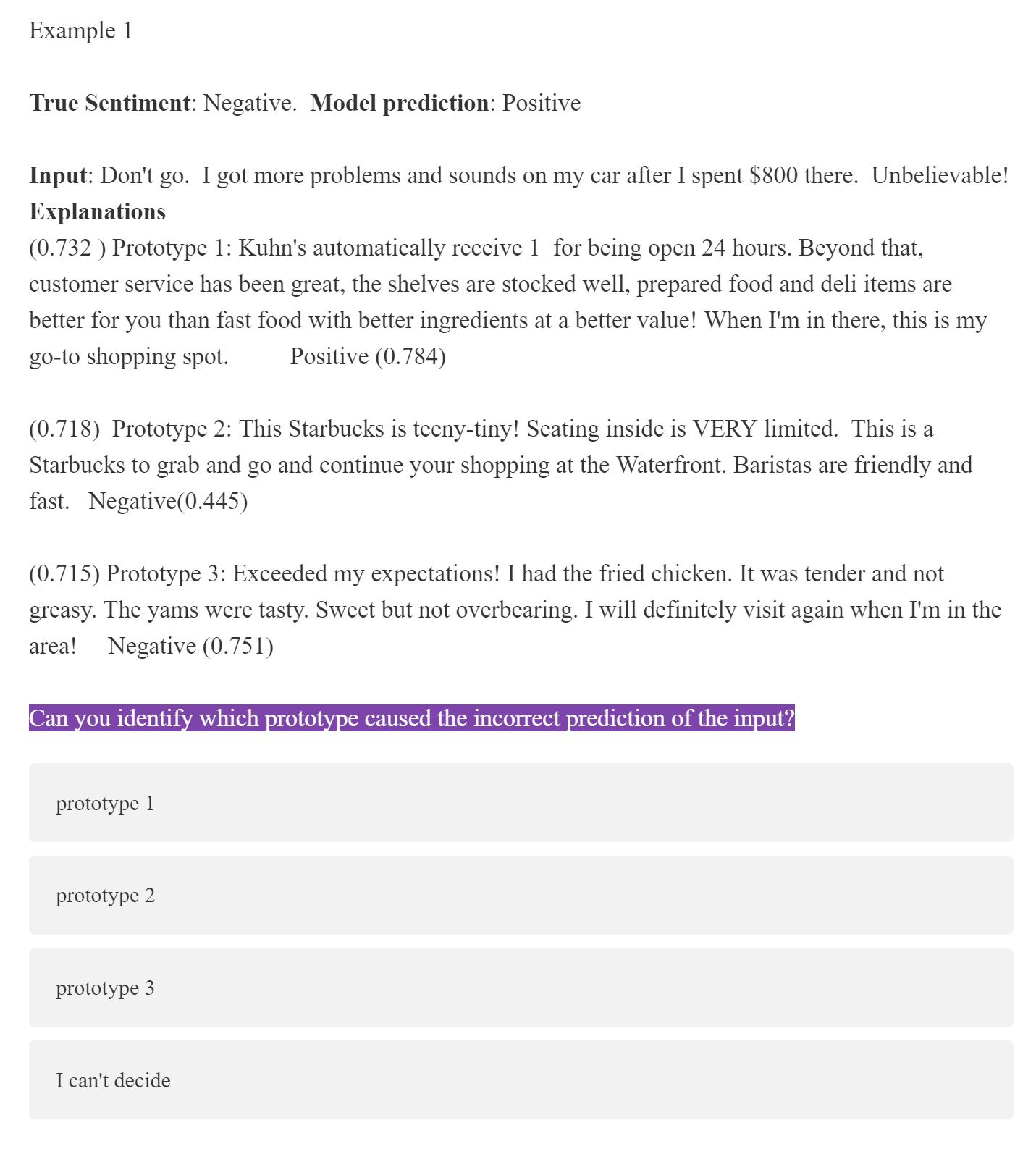}\caption{Diagnosis question for ProSeNet model.}\label{fig:survey_prosenet_diagnose}
\end{figure}

For diagnosing the ProSeNet and ProtoryNet, we create 3 questions for each model.
We show one example for each model in Figure \ref{fig:survey_ProtoryNet_diagnose} and Figure \ref{fig:survey_prosenet_diagnose}.

Figure \ref{fig:user_agreement} shows a sample question to ask if the users' choice matches the model's decision.
We trained 2 models with $K=15$ and $K=100$ prototypes and create 4 questions for each model; this gives us 8 questions total.

\begin{figure}[h!]
 \includegraphics[width=0.95\linewidth]{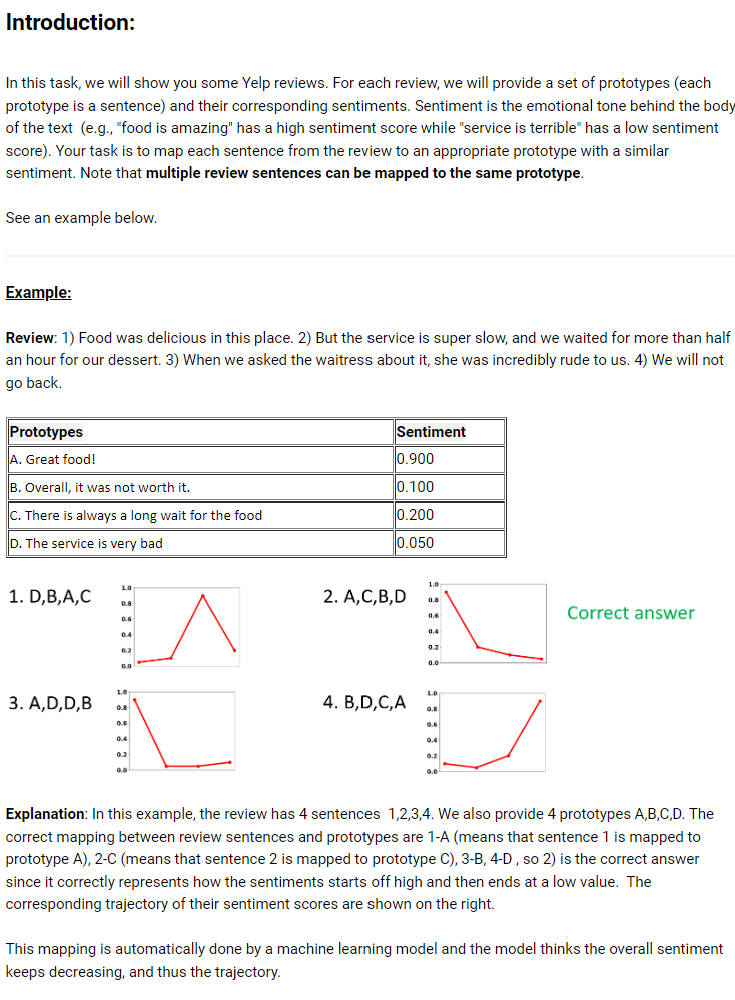}\caption{Prototype mapping question for ProtoryNet.}\label{fig:survey_trajectory}
\end{figure}

% \appendix 

\end{document}